\documentclass[runningheads]{llncs}

\usepackage{eccv}

\usepackage{graphicx}
\usepackage{subcaption}
\usepackage{float}
\usepackage[justification=justified]{caption}	%
\usepackage{lscape}                                         %

\usepackage[pagebackref,breaklinks,colorlinks,citecolor=eccvblue]{hyperref}

\usepackage{wrapfig}

\usepackage{booktabs}                   %
\usepackage{multirow}
\usepackage{makecell}

\usepackage{paralist}

\usepackage{bm}                          %
\usepackage{epsfig}
\usepackage{mathtools}
\usepackage{amssymb,amsmath}   %

\usepackage{units}
\usepackage{color}

\usepackage{comment}

\usepackage{url}  %

\usepackage[capitalize]{cleveref}
\crefname{section}{Sec.}{Secs.}
\Crefname{section}{Section}{Sections}
\Crefname{table}{Table}{Tables}
\crefname{table}{Tab.}{Tabs.}

\usepackage{xspace}
\usepackage[table]{xcolor}
\usepackage{setspace}

\usepackage{algorithm}
\definecolor{hcolor}{RGB}{238, 221, 240}
\usepackage{multirow}

\def\eg{e.g.,~}               %
\def\ie{i.e.,~}               %
\def\vs{vs.~}                 %

\newlength\paramarginsize
\newlength\figmarginsize
\newlength\secmarginsize
\newlength\figcapmarginsize
\newlength\tabcapmarginsize

\newcommand{\paramargin}{\vspace{\paramarginsize}}

\newcommand{\figcapmargin}{\vspace{\figcapmarginsize}}

\newcommand{\mpage}[2]
{
\begin{minipage}{#1\linewidth}\centering
#2
\end{minipage}
}

\newcommand{\topic}[1]
{
\paramargin\noindent \textbf{#1}
}

\newcommand{\figcaption}[2]
{
\caption{
\textbf{#1.}  %
#2            %
}
}

\newcommand{\secref}[1]{Section~\ref{sec:#1}}
\newcommand{\figref}[1]{Figure~\ref{fig:#1}} 
\newcommand{\tabref}[1]{Table~\ref{tab:#1}}
\newcommand{\eqnref}[1]{\eqref{eq:#1}}

\long\def\ignorethis#1{}

\def\ours{{\texttt{RCORE}}}
\def\cpr{{CPR}}
\def\torc{{TORC}}

\def\cg{{$\Delta_{\text{CG}}$}}
\def\mcg{{$\Delta_{\text{CG}}^{\text{macro}}$}}

\def\xi{\mathbf{x}_i}

\def\y{\mathbf{y}}
\def\f{\mathbf{f}}
\def\e{\mathbf{e}}

\def\s{\mathbf{s}}

\def\X{\mathbf{X}}

\def\F{\mathbf{F}}
\def\E{\mathbf{E}}
\def\M{\mathbf{M}}

\def\Acc{\mathrm{Acc}}
\def\Dcg{\Delta_\mathrm{CG}}

\graphicspath{{figure}, {images}, {example}}

\usepackage{eccvabbrv}

\usepackage{graphicx}
\usepackage{booktabs}

\usepackage[accsupp]{axessibility}  %

\usepackage{hyperref}

\usepackage{orcidlink}

\begin{document}

\title{Why Can’t I Open My Drawer? \\
Mitigating Object-Driven Shortcuts in Zero-Shot Compositional Action Recognition} 

\titlerunning{Why Can’t I Open My Drawer?}

\author{
Geo Ahn${}^{1}{}^{*}$\orcidlink{0009-0005-7792-2374} \and
Inwoong Lee${}^{2}$\orcidlink{0000-0003-4356-7616} \and
Taeoh Kim${}^{2}$\orcidlink{0000-0001-7252-5525} \and
Minho Shim${}^{2}$\orcidlink{0000-0002-9637-4909} \and
Dongyoon Wee${}^{2}$\orcidlink{0000-0003-0359-146X} \and
Jinwoo Choi${}^{1}{}^{\dagger}$\orcidlink{0000-0001-7043-0610}
}

\authorrunning{G. Ahn et al.}

\institute{%
${}^{1}$Kyung Hee University \quad ${}^{2}$NAVER Cloud \\
\email{\{ahngeo11, jinwoochoi\}@khu.ac.kr}\\
\email{\{inwoong.lee, taeoh.kim, minho.shim, dongyoon.wee\}@navercorp.com}
}

\maketitle
\let\thefootnote\relax\footnote{
${}^{*}$This work was done during an internship at NAVER Cloud.}
\let\thefootnote\relax\footnote{
${}^{\dagger}$Corresponding author.}
\begin{abstract}

Zero-Shot Compositional Action Recognition (ZS-CAR) requires recognizing novel verb--object combinations composed of previously observed primitives.
In this work, we tackle a key failure mode: models predict verbs via \emph{object-driven shortcuts} (\ie, relying on the labeled object class) rather than temporal evidence.
We argue that sparse compositional supervision and verb--object learning asymmetry can promote object-driven shortcut learning.
Our 
analysis with proposed diagnostic metrics
shows that existing methods overfit to training co-occurrence patterns and underuse temporal verb cues, resulting in weak generalization to unseen compositions.
To address object-driven shortcuts, we propose Robust COmpositional REpresentations (\ours{}) with two components.
Co-occurrence Prior Regularization (CPR) adds explicit supervision for unseen compositions and regularizes the model against frequent co-occurrence priors by treating them as hard negatives.
Temporal Order Regularization for Composition (TORC) enforces temporal-order sensitivity to learn temporally grounded verb representations.
Across Sth-com and EK100-com, 
\ours{} reduces shortcut diagnostics and consequently improves compositional generalization.
The code is available at \href{https://github.com/KHU-VLL/RCORE}{https://github.com/KHU-VLL/RCORE}.

\end{abstract}

\section{Introduction}
\label{sec:intro}

Many human actions can be decomposed into two semantic primitives—verbs and objects—and robust video understanding requires recognizing each component and reasoning over their interaction as a composition~\cite{goyal2017something, mat2020sthelse, damen2022ek100, diba2020hvu, bae2024devias}.
Zero-Shot Compositional Action Recognition (ZS-CAR) formalizes this goal: a model must recognize unseen verb–object pairs while keeping the verb and object vocabularies shared across splits~\cite{li2024c2c, jiang2025dhd, jung2025crr, ye2025logic}.

\begin{figure}[t]
    \centering
    \includegraphics[width=.98\linewidth]{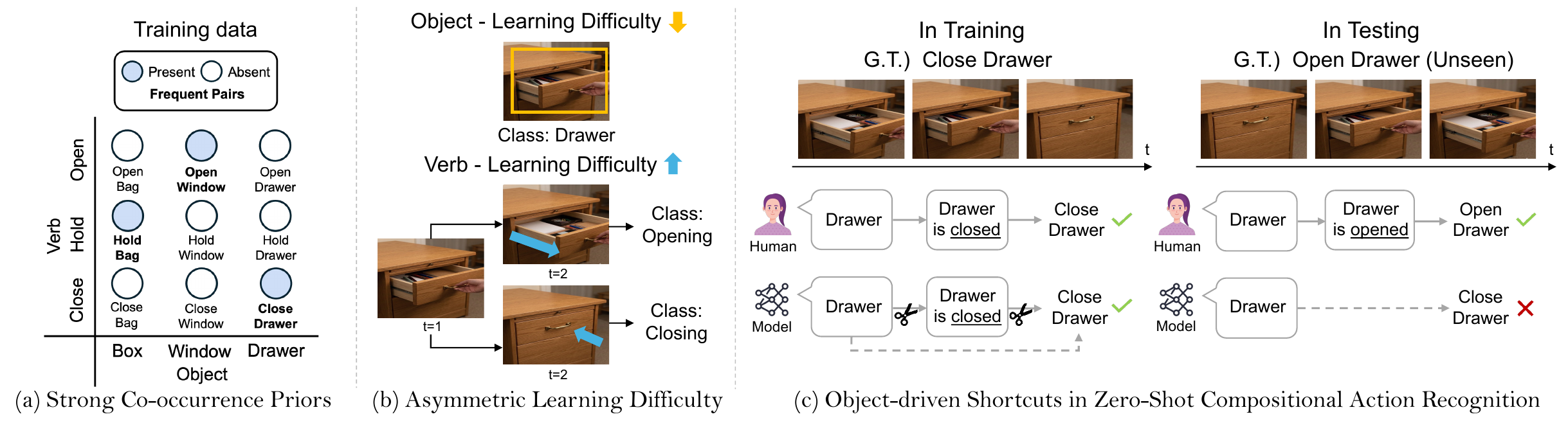}
    \vspace{-1em}
    \captionof{figure}{
    \textbf{Why object-driven shortcuts emerge in compositional video understanding?}
    (a) \textbf{Co-occurrence bias.} 
    Datasets are intrinsically sparse and highly skewed in their verb–object combinations, creating strong co-occurrence priors.
    (b) \textbf{Asymmetric learning difficulty.} 
    Objects are visually explicit and easy to recognize from a single frame, whereas verbs require multi-frame temporal reasoning.
    This imbalance makes object features dominate training signals.
    (c) \textbf{Object-driven shortcuts.} 
    Together, these two factors drive models to adopt \emph{object-driven shortcuts}, hindering genuine compositional generalization.
    Once the object is recognized, the model often predicts the most frequent verb paired with it, ignoring temporal evidence.
    }     
    \label{fig:teaser}
\end{figure}

In this setting, a key failure mode emerges: models often predict verbs via \emph{object-driven shortcuts} rather than temporal evidence.
Throughout, \emph{object} refers to the \emph{labeled noun class} in the verb--object vocabulary (not unlabeled background entities).
We argue that object-driven shortcuts arise from two issues inherent to ZS-CAR.
First, \textbf{compositional supervision is sparse and skewed}.
As illustrated in \figref{teaser} (a), datasets cover only a small fraction of the combinatorial verb--object space, and training labels concentrate on a limited set of frequent pairs~\cite{mat2020sthelse, damen2022ek100}.
This skew induces a strong \emph{co-occurrence prior} in the \emph{training data}—the empirical dominance of frequent verb–object pairs—which can be amplified into a \emph{model shortcut} at training time.
Under such skewed supervision, predictions for unseen compositions tend to drift toward \emph{seen} (especially frequent) pairs.
Second, \textbf{verbs are harder to learn than objects}.
As shown in \figref{teaser} (b), an object often becomes recognizable from a single frame, whereas a verb typically requires multi-frame temporal reasoning.
Prior work on shortcut learning suggests that models tend to rely on easier-to-learn cues rather than the intended evidence~\cite{nam2020lff,li2018resound,singh2020dont,geirhos2018cuecon,scimeca2022shortcut}.
In ZS-CAR, this verb--object difficulty gap makes object cues an especially attractive shortcut for verb prediction.

In this work, we posit that these two intertwined issues in verb-object compositional learning promote \emph{object-driven shortcuts}, which we diagnose as the key factor hindering ZS-CAR.
As \figref{teaser} (c) illustrates, a model can match a verb once it recognizes its frequently co-occurring object during training, without modeling the temporal dynamics essential to correctly recognize the verb.
This behavior weakens generalization to unseen compositions because it ties verb prediction to object-dependent priors rather than temporally grounded evidence.
We observe that models~\cite{li2024c2c}, even with a modern video-pretrained backbone~\cite{wang2024iv2}, indeed confuse verbs with opposite temporal order (\eg, opening \vs closing) on unseen compositions (see \secref{validation}).

To dissect how object-driven shortcuts hinder unseen generalization, we perform a comprehensive quantitative diagnosis. 
First, through controlled experiments, we show that the asymmetric learning difficulty between verbs and objects, together with skewed co-occurrence priors, promotes verb shortcut learning driven by object cues.
Next, we \emph{quantify} shortcut-driven failures in existing ZS-CAR models by two diagnostic ratios: False Seen Prediction (FSP) and False Co-occurrence Prediction (FCP), which measure how often an unseen input collapses to a seen—and especially frequent—training composition.
Using FSP/FCP, we show that a strong ZS-CAR baseline~\cite{li2024c2c} remains heavily biased toward seen compositions, resulting in poor generalization to unseen compositions.

Driven by the diagnosis, we propose Robust COmpositional REpresentations (\ours{}), targeting the two root causes of object-driven shortcuts.
First, Co-occurrence Prior Regularization (\cpr{}) expands supervision over originally absent compositions and suppresses the dominance of frequently seen pairs by treating them as hard negatives.
Second, Temporal Order Regularization for Composition (\torc{}) enforces temporal-order sensitivity so that the model learns temporally grounded verb representations instead of relying on static cues.

We validate \ours{} on Sth-com~\cite{li2024c2c} under a realistic open-world evaluation protocol that evaluates all possible verb-object pairs, without requiring test-set ground truth labels. 
We also introduce EK100-com, a ZS-CAR dataset repurposed from EPIC-KITCHENS-100~\cite{damen2022ek100}, which exhibits more severe compositional sparsity than Sth-com~\cite{li2024c2c}.
Across datasets and backbones, \ours{} reduces shortcut diagnostics (FSP/FCP) and improves unseen compositional generalization.

In this work, we make the following major contributions:
\begin{itemize}
    \item \textbf{Shortcut Diagnosis:} We provide a diagnostic view of \emph{object-driven shortcuts} in ZS-CAR. We empirically identify strong co-occurrence priors and asymmetric learning difficulty as the root causes of the shortcuts. Furthermore, we expose the limitations of baselines by quantifying how unseen inputs collapse into seen-especially frequent-training pairs, using FSP/FCP.
    \item \textbf{Diagnosis-Driven Framework:} We propose \ours{} comprising \cpr{} to suppress co-occurrence priors by expanding supervision over originally absent compositions and \torc{} to promote temporally grounded verb learning.
    \item \textbf{Effectiveness \& Robustness:} We demonstrate that \ours{} is effective, achieving superior performance on unseen compositions over baselines on Sth-com and EK100-com datasets, with multiple VLM backbones. 
    Moreover, through extensive quantitative analyses, we prove that our approach mitigates co-occurrence biases and learns temporally grounded verb features.
\end{itemize}

\section{Related Work}
\label{sec:related}

\topic{Compositional Action Recognition (CAR).}
CAR aims to recognize actions by factoring them into verbs and objects.
A recurring challenge is learning verb representations that are robust to static cues (\eg, objects/scenes) and capture temporal dynamics~\cite{mat2020sthelse, sun2021cdn, chatterjee2023oap,liu2025knowledgedriven}.
Prior work addresses this by modeling object-level interactions~\cite{mat2020sthelse, sun2021cdn} or by training verb representations to be less sensitive to object appearance~\cite{chatterjee2023oap}.
Related directions include \emph{open-vocabulary} CAR, which expands the action label space beyond a fixed taxonomy (often via text-driven semantics)~\cite{luo2022dark, chatterjee2023oap}.
In this work, we study Zero-Shot Compositional Action Recognition (ZS-CAR) under a fixed verb/object vocabulary, where verb and object label sets are shared across splits but some verb--object \emph{pairs are held out} for testing~\cite{li2024c2c, jiang2025dhd, jung2025crr, ye2025logic}.
In ZS-CAR, conditional learning~\cite{li2024c2c, jiang2025dhd} and disentanglement-oriented designs~\cite{jung2025crr} improve joint modeling, yet our study highlights that these models can still exhibit \emph{object-driven shortcut} behavior under sparse and skewed compositional supervision (see \secref{preliminaries}, \secref{validation}).

\topic{Compositional Zero-Shot Learning (CZSL) for image understanding.}
In the image domain, CZSL focuses on recognizing novel compositions of seen primitives (\eg, attribute–object) under sparse compositional supervision~\cite{misra2017red, naeem2021czsl, mancini2021compcos, kim2023hierarchical,wu2025cond, nayak2023csp}.
Representative approaches include feature disentanglement~\cite{saini2022disen, zhang2024disen}, modular factorization~\cite{purushwalkam2019taskdriven}, and CLIP-based prompt learning~\cite{nayak2023csp, li2024context}.
Compared to image CZSL, ZS-CAR additionally requires temporal grounding of verbs, which changes both the failure modes and the regularization opportunities.
In contrast, we reveal how sparse compositional supervision and the higher learning difficulty of verbs jointly amplify shortcut reliance in videos, and propose training-time regularizers tailored to ZS-CAR.

\topic{Shortcut learning.}
Models can learn \emph{shortcuts} by exploiting spurious correlations that are easier to fit than the intended evidence~\cite{geirhos2018cuecon,ilyas2019adversarial,nam2020lff,scimeca2022shortcut}.
Such shortcuts are closely tied to skewed training distributions and co-occurrence statistics, which can induce reliance on dataset priors~\cite{agrawal2018dont,singh2020dont}.
In compositional settings, prior work reweights rare combinations to mitigate co-occurrence bias~\cite{kim2023hierarchical}.
In video understanding, several works report reliance on static cues such as scenes~\cite{whycantchoi, wang2021be, ding2022fame, bae2024devias} and objects~\cite{bahng2019rebias, chatterjee2023oap} instead of motion-sensitive evidence.
Building on these observations, we study a ZS-CAR-specific manifestation: \emph{object-driven shortcuts} where object cues dominate verb prediction under sparse verb--object supervision, and we quantify this behavior with dedicated diagnostics.

\topic{Vision-Language Model (VLM).}
Recent VLMs broadly follow two paradigms: encoder-based models (\eg, CLIP~\cite{radford2021clip} and its video extensions~\cite{ma2022xclip,li2023unmasked,zhao2024videoprism,wang2024iv2}) 
and decoder-based models (\eg, LLaVA~\cite{liu2023llava}, Qwen-VL~\cite{bai2025qwen25vl}).
In ZS-CAR, however, inference and evaluation operate over a \emph{fixed} verb--object label space with \emph{explicit compositional scores} (\ie, logits over $\mathbb{Y}^V \times \mathbb{Y}^O$), which provides a controlled interface for diagnosis and for designing training-time regularizers that are \emph{transferable across backbones}.
Such decoder-based VLM interfaces often do not expose, in a standardized and reproducible way, explicit scores over the fixed label space, as they primarily provide language outputs at inference time.
Therefore, we focus on encoder-based VLMs in this controlled setting, and show that object-driven shortcuts persist even with a video-pretrained backbone~\cite{wang2024iv2}.

 \section{Diagnosis: Why ZS-CAR Models Fail?}
\label{sec:preliminaries}

\subsection{Zero-Shot Compositional Action Recognition}

We study Zero-Shot Compositional Action Recognition (ZS-CAR)~\cite{li2024c2c}, where the goal is to recognize \emph{unseen verb--object compositions} that are held out from the training set, while the verb and object vocabularies are \emph{fixed} and shared across train/validation/test splits.
Thus, unlike open-set or open-vocabulary settings, ZS-CAR does not introduce unseen \emph{verbs} or \emph{objects} at test time---only unseen \emph{compositions} (verb--object pairs).
Let $\mathbb{Y}^V$ and $\mathbb{Y}^O$ denote the sets of verb and object labels, and let each composition label be a pair $\y^C=(\y^V,\y^O) \in \mathbb{Y}^V \times \mathbb{Y}^O$.
The training set is $\mathbb{D}_{\text{train}}=\{(\X_i,\y^C_i)\}_{i=1}^N$, and its set of observed (seen) compositions is
$\mathbb{Y}_{\text{seen}}=\{\y^C_i \mid (\X_i,\y^C_i)\in\mathbb{D}_{\text{train}}\}$.
The space of unseen compositions is
$\mathbb{Y}_{\text{unseen}} = (\mathbb{Y}^V \times \mathbb{Y}^O)\setminus \mathbb{Y}_{\text{seen}}$.
The validation and test sets contain both seen and unseen compositions, enabling evaluation of compositional generalization.  

\begin{figure*}[t!]
\centering
\includegraphics[width=\linewidth]{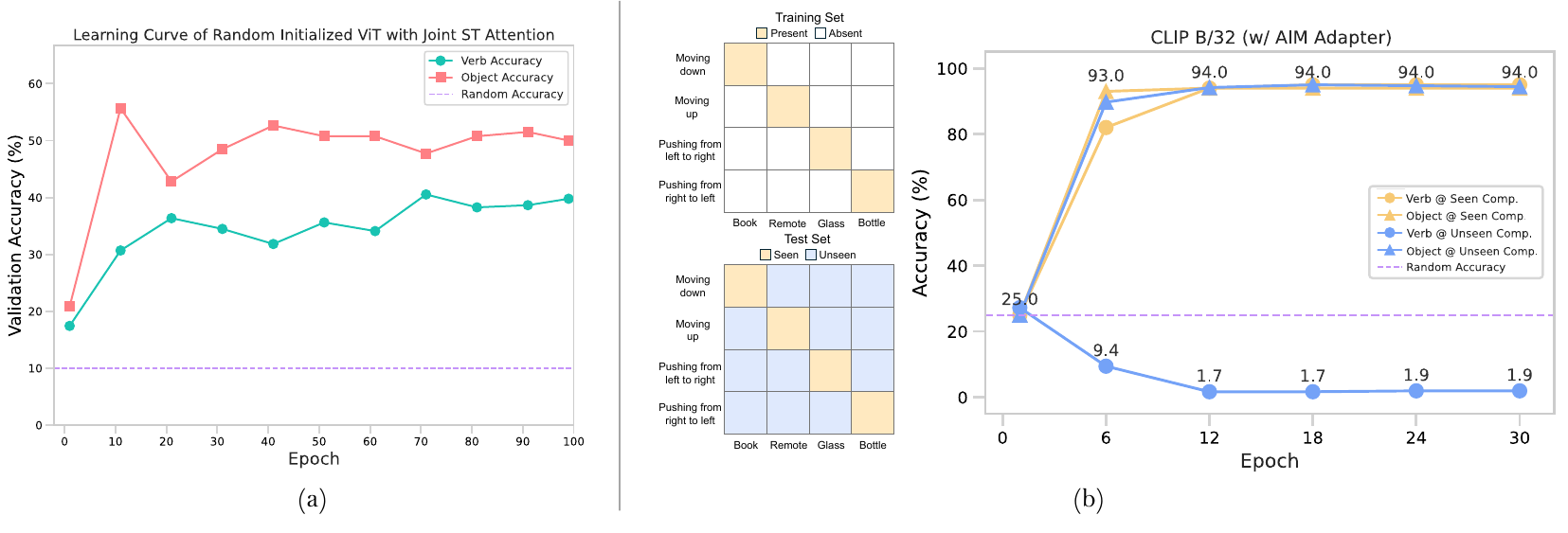}
\\
\figcapmargin
\vspace{-1em}
\figcaption{Controlled experiments demonstrate object-driven shortcut}{
    (a) \textbf{Objects are easier than verbs.} A randomly initialized ViT trained on a balanced $10{\times}10$ Sth-com subset learns objects substantially faster than verbs.
    (b) \textbf{Co-occurrence bias induces object-driven shortcuts.} On a perfectly biased training split, CLIP (with AIM~\cite{yang2023aim}) achieves high object accuracy, but verb accuracy on bias-conflict unseen compositions drops below chance, indicating shortcut-driven verb failures.
    Best viewed with zoom and color.
}
\label{fig:diagnosis_shortcut}
\end{figure*}

\subsection{Diagnostic Metrics}

To better understand ZS-CAR model behavior, we introduce two diagnostics:
(1) \emph{training-bias} metrics that quantify over-reliance on training co-occurrence statistics, and
(2) the \emph{Compositional Gap} that measures the benefit of modeling verb--object dependence beyond independent recognition.

\topic{Training-bias metrics.}
To quantify reliance on co-occurrence priors, we define two misclassification ratios on the \emph{unseen} subset of an evaluation split $\mathbb{Y}_{\text{unseen}}$.
\emph{False Seen Prediction (FSP)} is the fraction of misclassified unseen composition samples whose predictions $\hat{\y}^C$ fall into the seen composition categories $\mathbb{Y}_{\text{seen}}$.
\emph{False Co-occurrence Prediction (FCP)} is the fraction of misclassified unseen composition samples whose prediction $\hat{\y}^C$ falls into the frequent compositions $\mathbb{Y}_{\text{freq}}$.
Let $f(\y^C)$ denote the training frequency of a seen composition $\y^C \in \mathbb{Y}_{\text{seen}}$.
We define the set of frequent seen compositions $\mathbb{Y}_{\text{freq}}$ as
\begin{equation}
    \mathbb{Y}_{\text{freq}}=\{\y^C \in\mathbb{Y}_{\text{seen}} \mid f(\y^C) > \mu_f + \sigma_f\},    
    \label{eq:freq}
\end{equation}
where $\mu_f$ and $\sigma_f$ are the mean and standard deviation of $\{f(\y^C)\}_{\y^C\in\mathbb{Y}_{\text{seen}}}$.

To further analyze shortcut patterns, we decompose FSP/FCP into three component-wise cases based on prediction errors:
(i) Verb-collapse, where the predicted object is correct but the verb is incorrect;
(ii) Object-collapse, where the predicted verb is correct but the object is incorrect; and
(iii) Dual-collapse, where both verb and object are incorrect.
Using the metrics, we can identify whether co-occurrence bias primarily harms verb or object learning.

\topic{Compositional Gap.}
We introduce \textit{Compositional Gap} (\cg{}) to examine whether a model benefits from predicting the joint verb--object composition beyond getting the two components correct independently.
Given joint logits over $\mathbb{Y}^V \times \mathbb{Y}^O$, we obtain the predicted composition $\hat{\y}^C =(\hat{\y}^V,\hat{\y}^O)$ by selecting the top-1 composition label.
We then compute top-1 accuracies $\Acc^{V}$, $\Acc^{O}$, and $\Acc^{C}$ respectively.
Then we define \cg{} as
\begin{equation}
    \Delta_{\text{CG}} = \Acc^{C} - (\Acc^{V} \times \Acc^{O}).
    \label{eq:cg}
\end{equation}
We report \cg{} alongside FSP/FCP and unseen composition accuracy as a \emph{diagnostic} metric, rather than interpreting it as a standalone evaluation criterion.

\begin{figure}[t]
\centering
\mpage{0.3}{
\includegraphics[width=\linewidth]{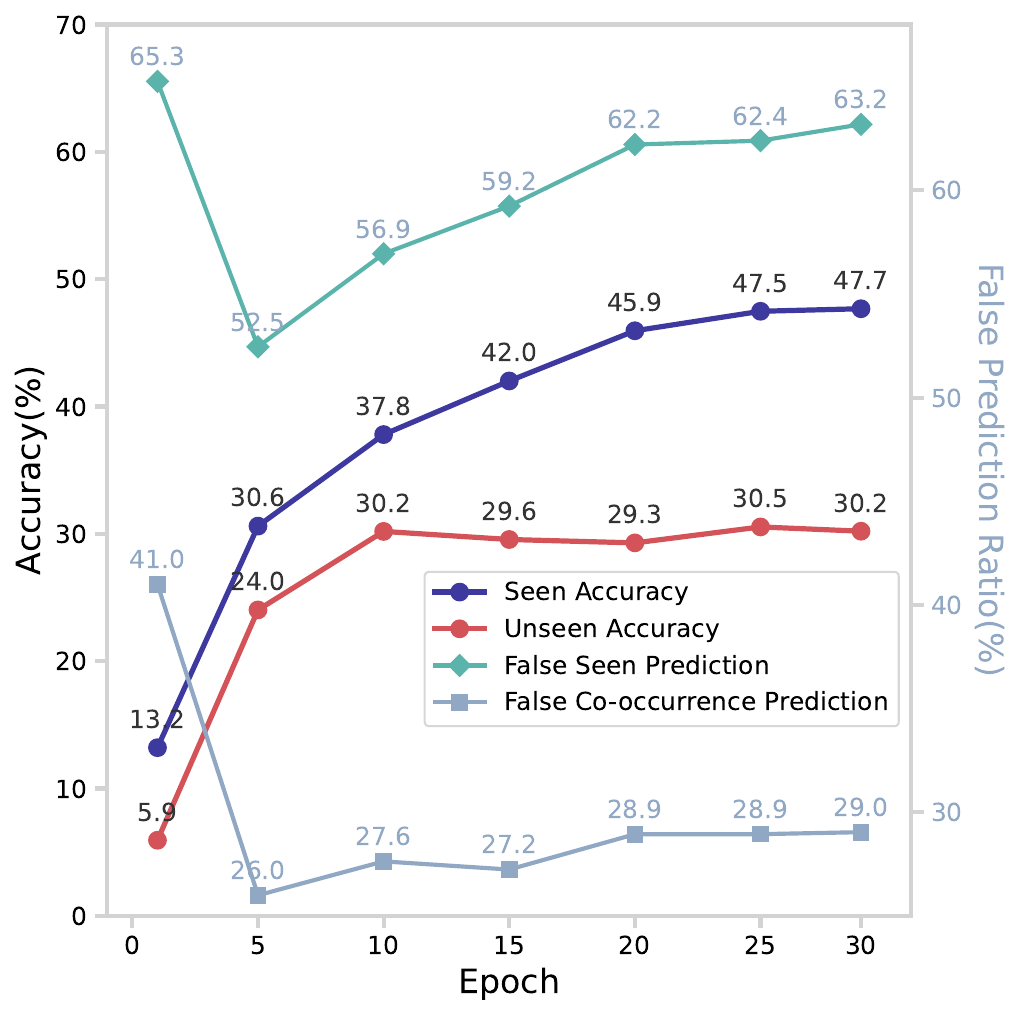}
}
\mpage{0.3}{
\includegraphics[width=\linewidth]{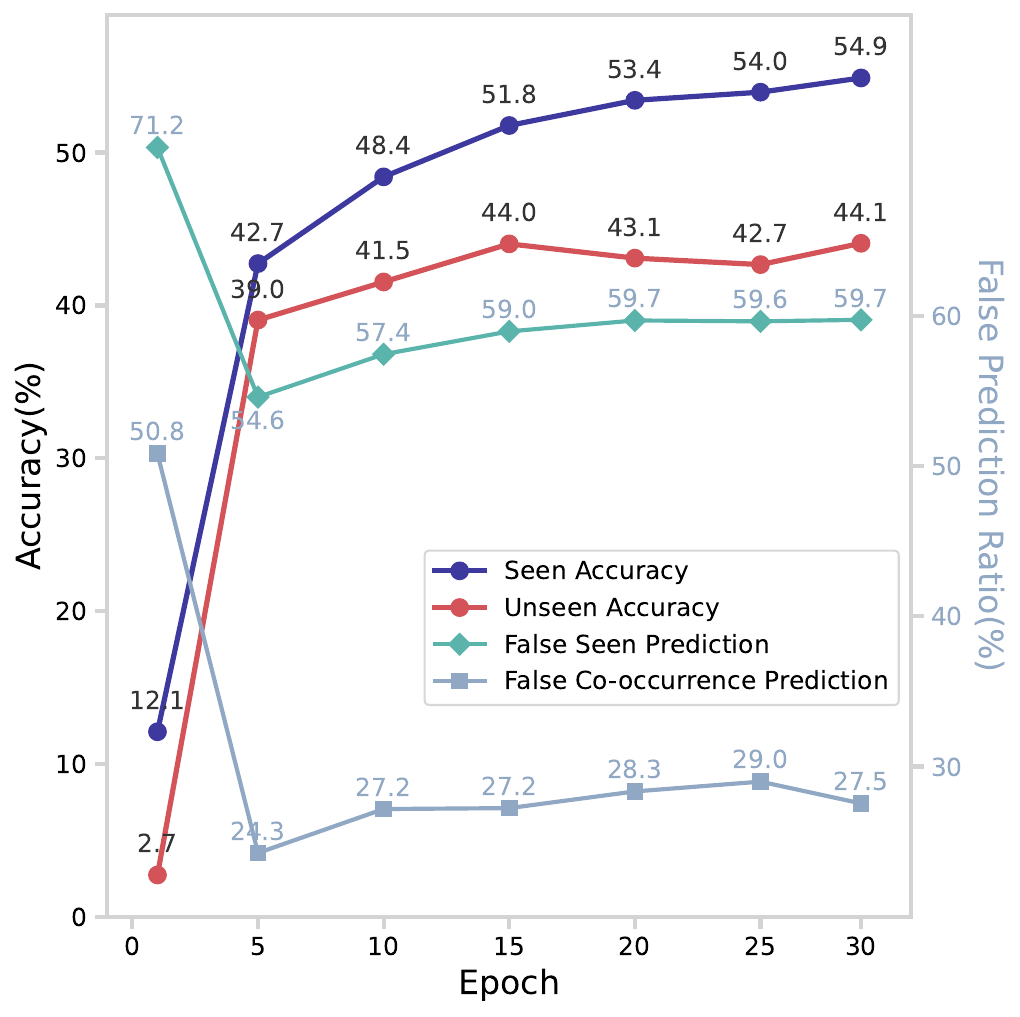}
}
\mpage{0.26}{
\includegraphics[width=\linewidth]{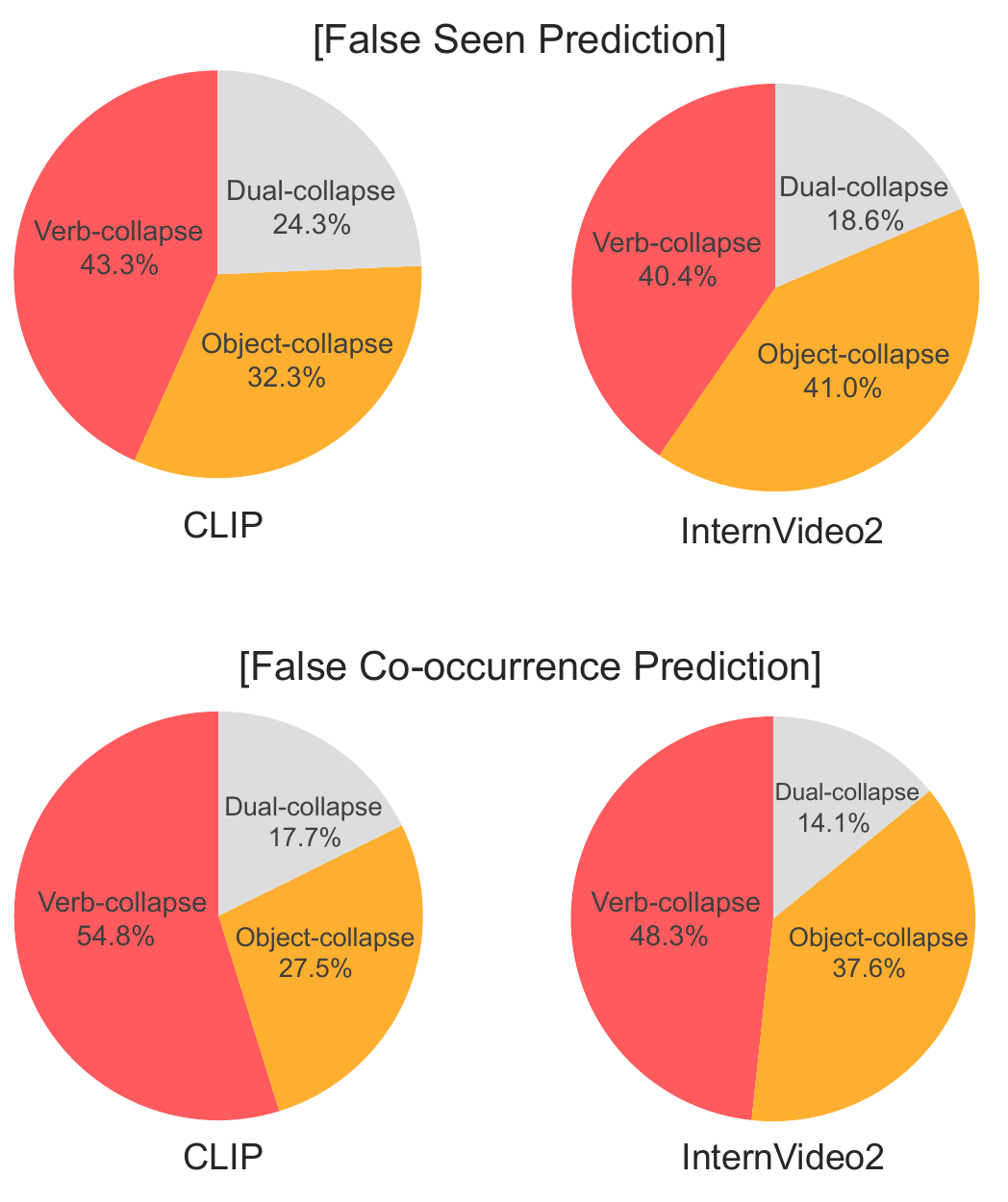}
}
\\
\vspace{-1em}
\mpage{0.3}{
\scriptsize{(a) CLIP}
}
\mpage{0.3}{
\scriptsize{(b) InternVideo2}
}
\mpage{0.26}{
\scriptsize{(c) FSP/FCP breakdown}
}
\\
\figcapmargin
\vspace{-.8em}
\figcaption{Shortcut reliance correlates with poor unseen generalization}{
We show learning curves of C2C~\cite{li2024c2c} on Sth-com with CLIP~\cite{radford2021clip} and InternVideo2~\cite{wang2024iv2}.
(a) With CLIP, the seen–unseen accuracy gap grows together with FSP/FCP, indicating strong overfitting to seen compositions.
(b) Even with the video-pretrained InternVideo2 backbone, the model still shows a notable seen–unseen gap and high FSP.
(c) FSP/FCP breakdown at the final epoch: CLIP is dominated by verb-collapse due to object-driven shortcuts, while InternVideo2 exhibits milder but persistent shortcut behavior.
Best viewed with zoom and color.
}

\label{fig:diagnosis_baseline}
\end{figure}

\subsection{Empirical Diagnosis of Object-driven Shortcuts in ZS-CAR}

In this section, we empirically diagnose object-driven shortcut behavior in ZS-CAR and analyze how it correlates with weak verb learning and poor generalization to unseen compositions.
We show that this behavior persists even with a modern video-pretrained VLM backbone~\cite{wang2024iv2}, and provide additional evidence in \secref{results} and \secref{additional_evidence}.

\topic{Objects are easier to learn than verbs.}
To compare the learning difficulty of verbs and objects, we train a randomly initialized ViT~\cite{dosovitskiy202vit} on a controlled subset of Sth-com~\cite{li2024c2c}.
Specifically, we construct a balanced $10\times10$ subset containing all 100 verb--object pairs to isolate learning dynamics from compositional sparsity.
Based on prior observations that models tend to fit easier cues earlier under limited supervision~\cite{nam2020lff, scimeca2022shortcut}, we compare the learning curves of verb \vs object prediction.
In \figref{diagnosis_shortcut} (a), the model learns objects faster and reaches higher accuracy than verbs under this controlled setting, suggesting that object prediction is easier than verb prediction in our ZS-CAR setup.

\topic{Object-driven shortcuts indeed exist in ZS-CAR.}
Under skewed co-occurrence statistics, models may rely on easier cues as shortcuts instead of learning the intended evidence~\cite{nam2020lff, scimeca2022shortcut, geirhos2018cuecon, bahng2019rebias, li2018resound, whycantchoi}.
We test whether CLIP exhibits such object-driven shortcuts by constructing a \emph{bias-controlled} training split where each verb is strongly correlated with a particular object (see construction details in \secref{imple_detail}), and evaluating on two splits:
(i) \emph{bias-aligned} seen compositions and (ii) \emph{bias-conflict} unseen compositions.
As shown in \figref{diagnosis_shortcut} (b), CLIP achieves high object accuracy \emph{on unseen compositions} while verb accuracy \emph{on unseen compositions} drops below the uniform-chance level ($1/|\mathbb{Y}^V|$), indicating that object cues dominate verb prediction under this biased supervision.
In \figref{shortcuts_vit}, we observe this trend even with random initialization, implying shortcuts stem from the training distribution and objective, not just pretraining.

\topic{Shortcut reliance correlates with poor unseen generalization.}
Next, we test whether shortcut reliance correlates with weak unseen generalization in a standard SOTA pipeline.
\figref{diagnosis_baseline} plots validation accuracies together with FSP/FCP for C2C~\cite{li2024c2c} on Sth-com~\cite{li2024c2c}.
As shown in \figref{diagnosis_baseline} (a), with a CLIP backbone~\cite{radford2021clip}, increasing FSP/FCP accompanies a widening seen--unseen gap, consistent with growing reliance on co-occurrence statistics.
Our error breakdown in \figref{diagnosis_baseline} (c) further shows that, among unseen samples that collapse to seen pairs, a large fraction corresponds to \emph{verb-collapse}, supporting the presence of object-driven shortcut behavior.
As shown in \figref{diagnosis_baseline} (b) and (c), with an InternVideo2~\cite{wang2024iv2} backbone, the trends are more stable; however, FSP remains high, and the FCP breakdown still exhibits substantial verb-collapse, suggesting that stronger video pre-training alone is insufficient to eliminate the tendency to overfit to seen pairs via co-occurrence-driven shortcuts.

\begin{wraptable}{r}{0.25\textwidth}
\centering
\vspace{-4.5em}
\caption{\textbf{$\Dcg{}$ on Sth-com.}
}
\resizebox{\linewidth}{!}{
\begin{tabular}{l l c}
\toprule
Backbone & Split & $\Delta_{\text{CG}}$ \\
\midrule
\multirow{2}{*}{CLIP} 
& Seen & +3.24 \\
& Unseen & $-$0.42 \\
\midrule
\multirow{2}{*}{InternVideo2} 
& Seen & +2.24 \\
& Unseen & $-$0.63 \\
\bottomrule
\end{tabular}
}
\label{tab:cg_table}
\vspace{-2em}
\end{wraptable}

\topic{\cg{} indicates degraded compositional behavior on unseen compositions.}
We report \cg{} of C2C~\cite{li2024c2c} with CLIP~\cite{radford2021clip} and InternVideo2~\cite{wang2024iv2} backbones.
As shown in \tabref{cg_table}, the model exhibits negative \cg{} on unseen compositions even with the video-pretrained backbone, indicating that $\Acc^{C}$ falls below the independent reference $\Acc^{V}\times\Acc^{O}$.
Together with the training-bias metrics (FSP/FCP), these results support that shortcut reliance is closely associated with weak unseen compositional generalization in current ZS-CAR pipelines.

\section{\ours{}}
\label{sec:method}

We introduce \ours{}, a diagnosis-driven learning framework that mitigates \emph{object-driven shortcuts} in ZS-CAR.  
\ours{} improves compositional generalization by strengthening verb–object representations under sparse supervision through two complementary components:  
(i) Co-occurrence Prior Regularization (\cpr{}) expands supervision over novel compositions and leverages frequently seen pairs as hard negatives to suppress co-occurrence priors, and  
(ii) Temporal Order Regularization for Composition (\torc{}) loss promotes temporally grounded verb learning over static object cues.
Following prior work~\cite{li2024c2c, jung2025crr}, we employ an adapter-based tuning, 
AIM~\cite{yang2023aim} for CLIP~\cite{radford2021clip} and LoRA~\cite{hu2022lora} for InternVideo2~\cite{wang2024iv2}, as our backbone.  
\figref{overview} (a) gives an overview of \ours{}.

\topic{Feature extraction.} 
Given a video $\X$ with $T$ frames, the backbone encoder outputs frame-level features $\F \in \mathbb{R}^{T \times D}$.  
We transform them into verb features $\F^V \in \mathbb{R}^{T \times D}$ and object features $\f^O \in \mathbb{R}^D$ using dedicated encoders.  
We obtain text embeddings for verbs and objects, $\E^V \in \mathbb{R}^{|\mathbb{Y}^V| \times D}$ and $\E^O \in \mathbb{R}^{|\mathbb{Y}^O| \times D}$, by feeding class-specific prompts 
into the text encoder.
\vspace{-1.0em}

\subsection{CPR: Co-occurrence Prior Regularization}
\label{sec:cpr}

To mitigate object-driven shortcuts under sparse and skewed compositional supervision, we propose \cpr{}, a training strategy that injects supervision for \emph{synthesized} verb--object pairs while preserving the stability of closed-world optimization.
\cpr{} combines (i) synthesized composition supervision, (ii) a margin-based regularizer that down-weights \emph{frequently} seen hard negatives, and (iii) a batch-adaptive expansion of the composition label space to make these new pairs trainable without optimizing over the full $\mathbb{Y}^V \times \mathbb{Y}^O$ space.

\begin{figure*}[t]
\centering
\includegraphics[width=\linewidth]{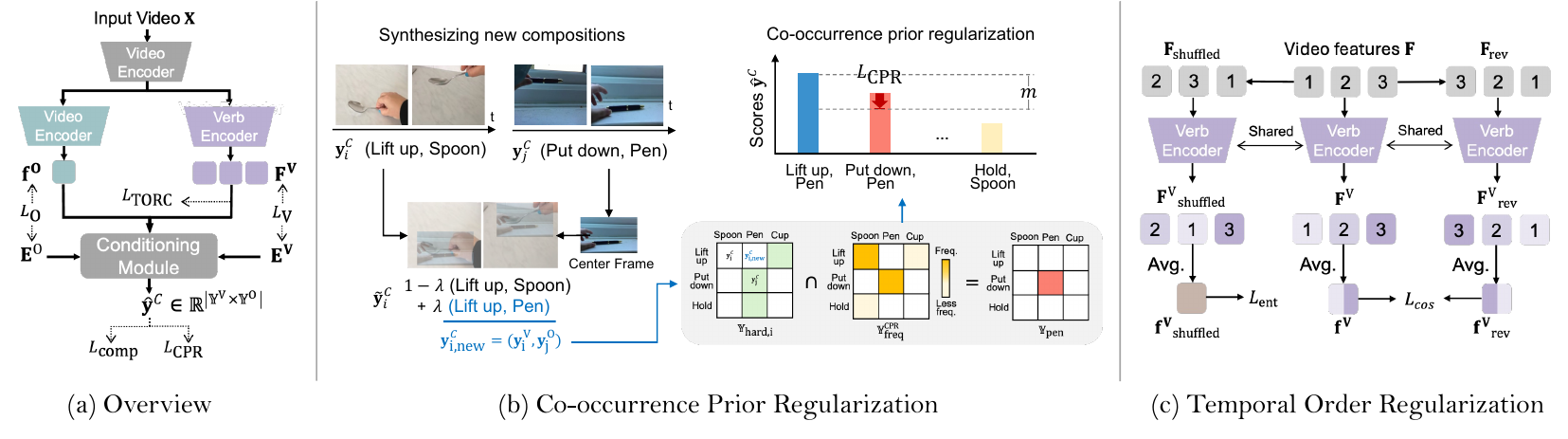} \\
\hfill
\hfill
\vspace{-2em}
\figcaption{Overview of \ours{}}{
(a) Overview of our proposed \ours{} framework.
(b) \cpr{} synthesizes plausible yet unseen verb–object compositions and penalizes frequently seen hard negatives by a margin-based regularizer.
(c) \torc{} penalizes alignment between original and temporally perturbed feature vectors, enforcing explicit temporal order modeling and reducing object-driven shortcuts.
}
\label{fig:overview}
\end{figure*}

\topic{Synthesized composition supervision.}
To provide explicit supervision for compositions absent from the training set, we first synthesize videos that represent new \emph{composition} labels.
Given a training sample $\X_i$ with composition label $\y_i^C=(\y_i^V,\y_i^O)$, we construct a new video $\tilde{\X}_i$ by injecting the static object cue from another video $\X_j$ (with $\y_j^O \neq \y_i^O$) into the high-motion regions of $\X_i$, as shown in \figref{overview} (b).
Specifically, for each frame $k\in\{1,\ldots,T\}$ we obtain
\begin{align}
\tilde{\X}_i^{k}
&= \bigl(\mathbf{1}-\lambda\,\M_i^{k} \bigr)\odot \X_i^{k}
\;+\;
\bigl(\lambda\,\M_i^{k}\bigr)\odot \X_j^{\lfloor T/2 \rfloor},
\label{eq:voca}
\end{align}
where $\lambda\in[0,1]$ controls the injection strength and $\M_i^{(k)}\in\{0,1\}^{H\times W}$ is the high-motion region mask for frame $k$, extracted by a learning-free estimator~\cite{ding2022fame}.
The operator $\odot$ denotes element-wise multiplication, with $\M_i^{k}$ broadcast across channels.
Accordingly, we use soft labels for the synthesized video as follows:
\begin{align}
\tilde{\y}^O_i &= (1-\lambda)\,\y_i^O + \lambda\,\y_j^O, \\
\tilde{\y}^C_i &= (1-\lambda)\,\phi(\y_i^V,\y_i^O) + \lambda\,\phi(\y_i^V,\y_j^O),
\label{eq:voca_label}
\end{align}
where $\phi:\mathbb{Y}^V \times \mathbb{Y}^O \rightarrow \{0,1\}^{|\mathbb{Y}^V \times \mathbb{Y}^O|}$ maps a composition $(\y^V,\y^O)$ to its one-hot vector in the flattened joint label space.

\topic{Co-occurrence prior regularization loss.}
Even after adding supervision for synthesized compositions, skewed co-occurrence statistics can still make the model over-score a small set of \emph{frequently} seen pairs.
To counter this score collapse, we enforce a margin constraint that makes the \emph{target synthesized} composition
$\y^C_{i,\text{new}}=(\y_i^V,\y_j^O)$
outrank \emph{frequent seen hard negatives} by at least $m$.
We define the hard-negative label set for $\y^C_{i,\text{new}}$ as follows:
\begin{align}
\mathbb{Y}_{\text{hard},i}
=
\left\{
(\y^V,\y^O)~\middle|~\big[(\y^V=\y_i^V)\oplus(\y^O=\y_j^O)\big] \land (\y^V,\y^O)\neq(\y_i^V,\y_i^O)
\right\},
\label{eq:hard}
\end{align}
where $\oplus$ denotes the exclusive or operator.
Let $\mathbb{Y}_{\text{freq}}^\text{CPR} \subset \mathbb{Y}_{\text{seen}}$ be the set of frequently seen compositions 
(see \secref{imple_detail_ours} for the exact criterion).
Then we define the CPR loss with margin $m$ over the penalty set $\mathbb{Y}_{\text{pen}}=\mathbb{Y}_{\text{hard},i} \cap \mathbb{Y}_{\text{freq}}^\text{CPR}$ as:\\
\begin{align}
L_{\text{CPR}}
= \sum_{\y^C \in \mathbb{Y}_{\text{pen}}}
\max\Big(0,\; s(\y^C) - s(\y^C_{i,\text{new}}) + m\Big),
\label{eq:cpr_margin}
\end{align}
\\
where $s(\y^C)$ denotes the logit assigned to the composition label $\y^C$.

\topic{Batch-adaptive label-space expansion.}
Conventional ZS-CAR pipelines optimize classification only over \emph{seen} compositions $\mathbb{Y}_{\text{seen}}$ (closed-world over compositions), which makes it non-trivial to directly supervise \emph{novel} verb--object pairs during training.
On the other hand, naively applying cross-entropy over the full $\mathbb{Y}^V \times \mathbb{Y}^O$ space can harm unseen accuracy by treating most unseen compositions as negatives throughout training (see \secref{ablation}).
To provide an effective supervision for novel compositions, we introduce a \emph{batch-adaptive label space expansion} strategy.
For each mini-batch, we construct an expanded label set as $\mathbb{Y}_{\text{exp}} = \mathbb{Y}_{\text{seen}} \cup \mathbb{Y}_{\text{new}}$, where $\mathbb{Y}_{\text{new}}$ denotes the label set for the newly synthesized videos.
We compute cross-entropy only over $\mathbb{Y}_{\text{exp}}$.
This preserves the optimization stability of closed-world training while progressively injecting supervision for new compositions.

\subsection{TORC: Temporal Order Regularization for Composition}

To counter object-driven shortcuts in verb learning, we introduce \torc{}, a temporal \emph{order-sensitivity} regularizer.
While \cpr{} expands supervision over compositions, verb recognition can still collapse to static object cues due to the asymmetric learning difficulty between verbs and objects.
As shown in \figref{overview} (c), \torc{} directly regularizes the verb representation to depend on temporal structure: it (i) discourages invariance to temporal reversal by separating forward and reversed features, and (ii) penalizes over-confident verb predictions when temporal order is disrupted by encouraging high-entropy outputs.
Together, these constraints reduce reliance on static cues and promote temporally grounded verb representations.

\topic{Temporal perturbation for regularization.}
Given frame-level features $\F = (\f_1,\dots,\f_T)$, we form two regularization views:  
(i) a reversed sequence $\F_{\text{rev}} = (\f_T,\dots,\f_1)$, and  
(ii) a temporally shuffled sequence $\F_{\text{shuffled}} = \pi(\F)$, where $\pi$ is a random permutation sampled from $\mathcal{P}$.  
We then feed $\F_{\text{rev}}$ and $\F_{\text{shuffled}}$ into the verb encoder to obtain perturbed verb features $\f^V_\text{rev}\in \mathbb{R}^{D}$ and $\f^V_\text{shuffled} \in \mathbb{R}^{D}$.

\topic{\torc{} loss.}
We first push the model to distinguish forward and reversed temporal semantics.  
We minimize the cosine similarity between the original verb $\f^V$ and the reversed feature vectors $\f^V_{\text{rev}}$: $L_{\cos} = \frac{\mathbf{f^V}^\top\f^V_{\text{rev}}}{\|\f^V\|\|\f^V_{\text{rev}}\|}$.
Temporal reversal often flips action meaning (\eg, opening \vs closing), yet the state-of-the-art model~\cite{li2024c2c} yields a high cosine similarity ($0.92$; see \figref{cosine_sim}), indicating weak temporal discrimination.

Next, we regularize the model to avoid confident verb predictions when temporal structure is disrupted.  
Given the verb text embeddings $\mathbf{E}^V = \{\e^V_m\}_{m=1}^{|\mathbb{Y}^V|}$, we define the negative entropy loss $L_\text{ent} = \sum_{m=1}^{|\mathbb{Y}^V|} p_m \log p_m$ with the predicted probability $p_m$ of the $m$-th verb class based on the cosine similarity between visual and textual features as:
\begin{align}
    p_m &= \frac{\exp(\cos(\f^V_\text{shuffled}, \e^V_m) / \tau)}{\sum_{n=1}^{|\mathbb{Y}^V|} \exp(\cos(\f^V_\text{shuffled}, \e^V_n) / \tau)},
\label{eq:ent}
\end{align}
where $\tau$ is a temperature parameter.
Maximizing entropy prevents the model from inferring verbs from static spatial cues alone, enforcing reliance on temporal evidence.
We define the \torc{} loss as: $L_{\text{TORC}} = L_{\cos} + L_{\text{ent}}$.

\subsection{Training}
\label{sec:training_loss}

\noindent\textbf{Component loss.}
We compute verb logits $\mathbf{S}_V \in \mathbb{R}^{|\mathbb{Y}^V|}$ and object logits $\mathbf{S}_O \in \mathbb{R}^{|\mathbb{Y}^O|}$ based on the cosine similarity scores.  
We apply standard cross-entropy losses for verb and object prediction: $L_{\text{com}} = L_V + L_O$.

\noindent\textbf{Composition loss.}
Following prior work~\cite{li2024c2c}, we obtain the composition logits $\mathbf{S}_C \in \mathbb{R}^{|\mathbb{Y}^V \times \mathbb{Y}^O|}$ by aggregating factorized logits: $\mathbf{S}_C = \mathbf{S}_V \boxplus \mathbf{S}_{O|V} + \mathbf{S}_O \boxplus \mathbf{S}_{V|O}$, where $\mathbf{S}_{O|V}$ and $\mathbf{S}_{V|O}$ represent the conditional logits and $\boxplus$ denotes broadcasted addition into the joint verb--object space.
Then we employ the cross-entropy loss integrated with our batch-adaptive label space expansion (\secref{cpr}). 
Given $\s_c \in \mathbf{S}_C$, the final composition loss utilizing the dynamically constructed denominator $\mathbb{Y}_{\text{exp}}$ is formulated as: $L_{\text{comp}}
= - \log (\exp(\hat{s}_{c})/\sum_{c=1}^{|\mathbb{Y}_{\text{exp}}|} \exp(s_c))$.

\noindent\textbf{Total loss.}
We train \ours{} with the combined objective $L_{\text{total}}$ defined as
$L_{\text{total}}
= \alpha L_{\text{com}}
+ \beta L_{\text{comp}}
+ \gamma L_{\text{TORC}}
+ \delta L_{\text{CPR}},$
where $\alpha$, $\beta$, $\gamma$ and $\delta$ are hyperparameters.

\section{Experimental Results}
\label{sec:results}

Building on our diagnosis, we evaluate whether \ours{} reduces shortcut reliance and improves compositional generalization.
We first describe the experimental setup (\secref{exp_setup}), then present diagnostic analyses (\secref{validation}), followed by quantitative comparisons (\secref{main_results}), and finally ablation studies (\secref{ablation}).

\subsection{Experimental setup}
\label{sec:exp_setup}

\topic{Limitations of conventional evaluation.}
Existing ZS-CAR works \cite{li2024c2c, jiang2025dhd, jung2025crr} rely on the \emph{closed-world} protocol, where inference is restricted to compositions that appear in the validation/test split.  
This protocol hides a model’s tendency to over-predict seen compositions and obscures co-occurrence-driven behaviors.
More critically, \emph{test-set–tuned} bias calibration selects a bias term using test labels, inflating unseen accuracy and undermining fair comparison.

\topic{Our evaluation setup.}
To expose genuine generalization, we adopt an \emph{open-world, unbiased} protocol by default:
(i) inference spans the full space $\mathbb{Y}^V \times \mathbb{Y}^O$, and  
(ii) no test labels are used to tune biases.  
We also report closed-world H.M. and AUC for comparison with existing methods.
We show open-world biased results using a validation set-tuned bias in \tabref{biased_sth_results}.

\topic{Datasets.}
We evaluate on two ZS-CAR benchmarks: Sth-com~\cite{li2024c2c} and our curated EK100-com.  
Sth-com is derived from Something-Something V2~\cite{goyal2017something} and Something-Else~\cite{mat2020sthelse},  
containing 79K videos with 161 verbs and 248 objects.  
Following prior work~\cite{yun2022time,laura2021temporal}, we also evaluate on the \emph{Temporal} subset consisting of verbs requiring temporal reasoning.
We additionally introduce the \textbf{EPIC-KITCHENS-100-composition (EK100-com)} dataset, constructed by repurposing EK100~\cite{damen2022ek100} following the Sth-com protocol.  
EK100-com includes 71K egocentric videos, 81 verbs, and 216 objects.  
Due to the long-tailed label distribution of EK100, it exhibits \emph{severe compositional sparsity}.
Its label coverage ratio
is only $7.5\%$, which is considerably lower than the $12.8\%$ observed in Sth-com~\cite{li2024c2c}.
Full details are in \secref{ek100-com}.

\begin{figure}[t]
\centering    
\mpage{0.27}{
\centering    
\includegraphics[width=\linewidth]{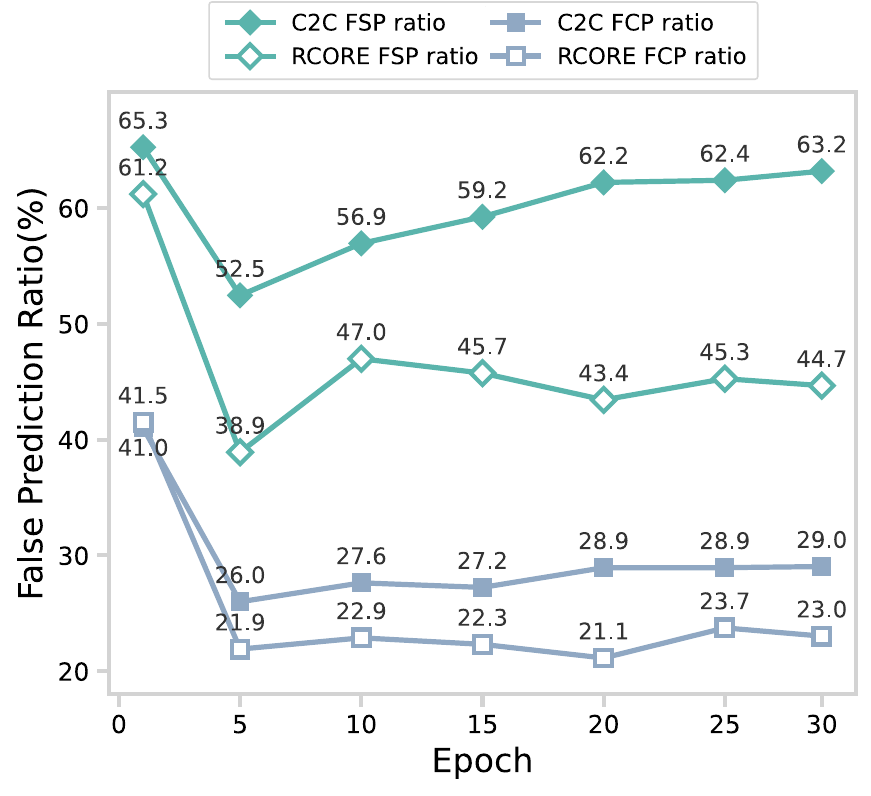}
\\
\scriptsize{\hspace{.5cm}(a)}
} 
\mpage{0.27}{
\centering    
\includegraphics[width=\linewidth]{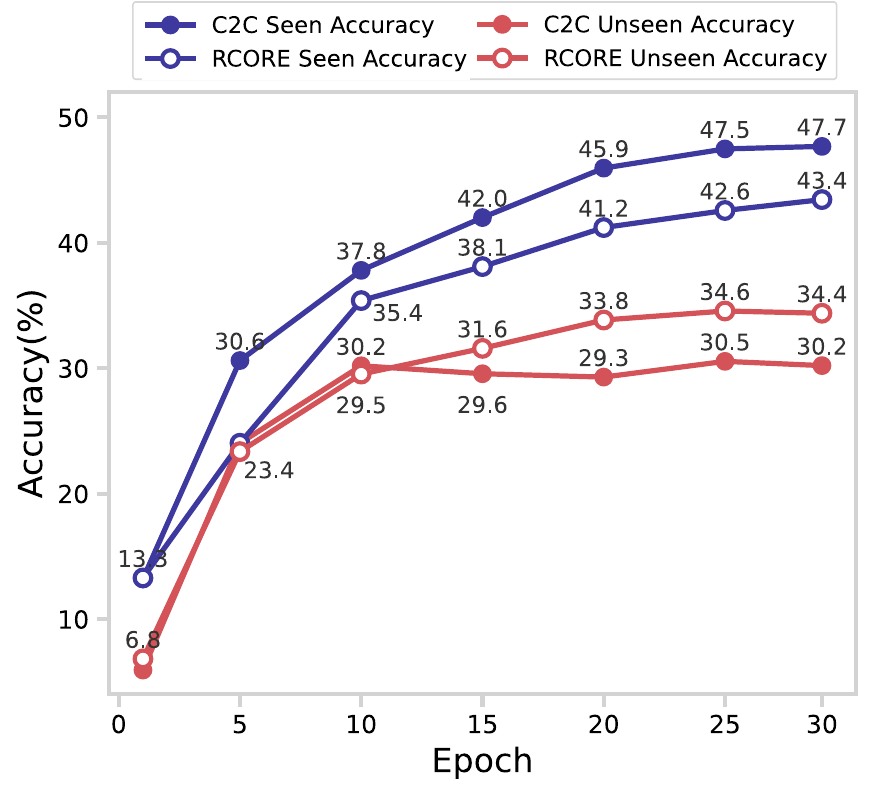}
\\
\scriptsize{\hspace{.5cm}(b)}
}
\mpage{0.33}{
\centering    
\includegraphics[width=\linewidth]{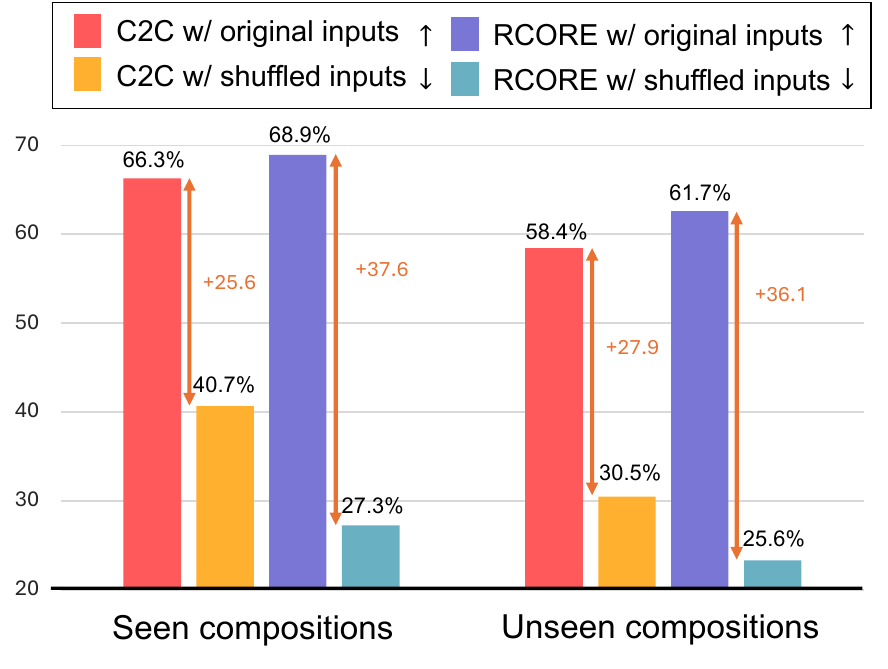}
\\
\scriptsize{\hspace{.5cm}(c)}
} 
\vspace{-1em}
\figcapmargin
\figcaption{Analysis on the effects of \ours{} on the Sth-com~\cite{li2024c2c} dataset}{
    (a) \ours{} suppresses the growth of FSP/FCP during training, unlike the baseline.
    (b) This reduces the seen–unseen gap and improves unseen composition accuracy.
    (c) On the temporal subset, \ours{} shows a larger drop under temporal shuffling, indicating stronger temporal dependence over static cues.
    Best viewed with zoom and color.
}
\label{fig:validation}
\end{figure}

\topic{Evaluation metrics.}
We report top-1 composition, verb and object accuracies.
Verb and object accuracies are conditioned on the \emph{seen/unseen composition}: \emph{verb@seen-comp} and \emph{object@seen-comp} are computed on samples with $\y^C \in \mathbb{Y}_{\text{seen}}$, while \emph{verb@unseen-comp} and \emph{object@unseen-comp} are computed on samples with $\y^C \in \mathbb{Y}_{\text{unseen}}$.
We report the corresponding harmonic mean (H.M.) between seen-comp and unseen-comp accuracies for composition, verb, and object.
We also report the compositional gap $\Dcg$~\eqnref{cg}.
For a detailed illustration of our evaluation setting, please refer to \secref{zscar_setting}.

\topic{Baselines.}
We use two encoder-based VLMs, CLIP-B/16~\cite{radford2021clip} and InternVideo2-CLIP-B/14~\cite{wang2024iv2}, as backbones for ZS-CAR methods.
We also report the InternVi-deo2-1B\cite{wang2024iv2} backbone results in \secref{iv2_1b_results}.
We primarily compare \ours{} against the current SOTA, C2C~\cite{li2024c2c} and Jung et al.~\cite{jung2025crr}; since the code for the latter's CLIP variant is not publicly available, we report our re-implementation that reproduces its closed-world performance.
We also compare against LogicCAR~\cite{ye2025logic}, a recent ZS-CAR method, in the closed-world setting only, as its code is not publicly available.
Since it builds on C2C~(enhance)~\cite{li2024c2c}, a variant of C2C~\cite{li2024c2c} with additional training constraints,  we apply \ours{} to the same enhanced backbone (denoted as \ours{}~(enhance)) and report its results for a fair comparison.
To isolate the effect of conditional learning, we further evaluate independent modeling baselines, `AIM'~\cite{yang2023aim} (for CLIP) and `LoRA'~\cite{hu2022lora} (for InternVideo2), named after their respective adapter tuning methods.

\subsection{Analysis}
\label{sec:validation}

We empirically show that \ours{} reduces reliance on co-occurrence priors and mitigates object-driven shortcuts in verb learning, yielding more robust compositional behavior than the baseline~\cite{li2024c2c}.

\begin{wrapfigure}{r}{0.5\textwidth}
    \vspace{-3.5em}
    \centering
    \includegraphics[width=\linewidth]{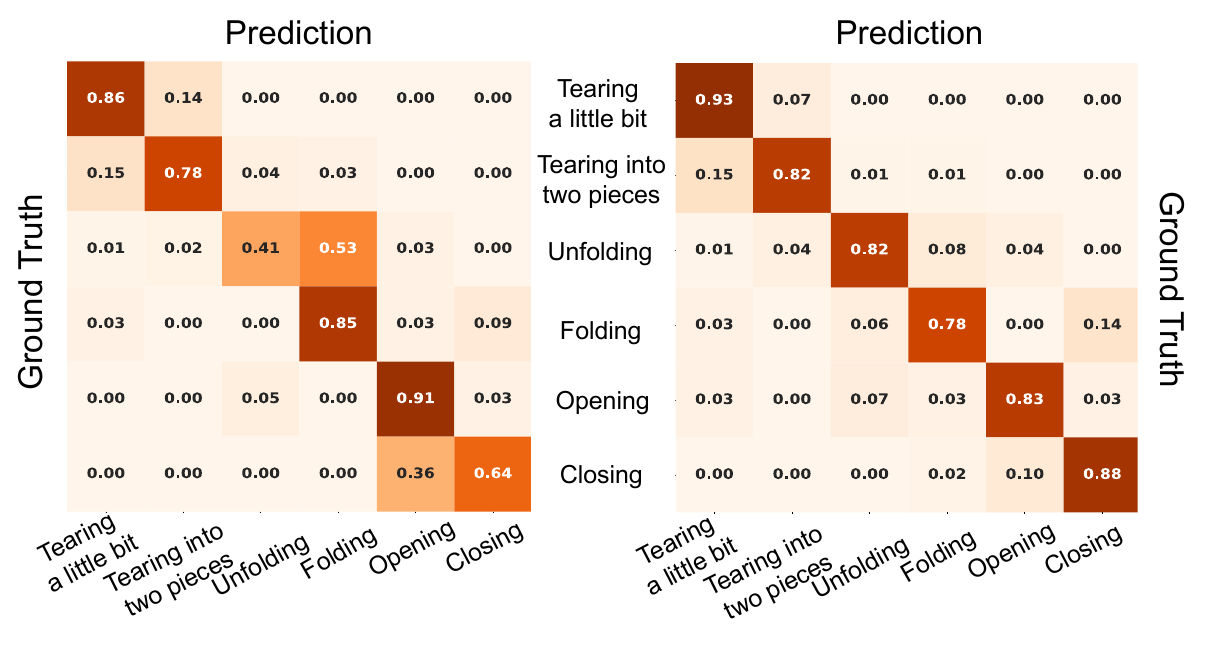}
    \\
    \vspace{-1em}
    {\fontsize{7pt}{10pt}\selectfont 
    C2C (Baseline)
    \centering \hspace{1.2cm} 
    \ours{} (Ours)
    }
    \\
    \vspace{-1em}
    \figcaption{\ours{} mitigates object-driven shortcuts in verb learning}
    {
        We visualize confusion matrices for six representative verbs on \emph{unseen} compositions of the Sth-com~\cite{li2024c2c} test set.
        All values are normalized frequencies across the six verb classes.
    }
    \vspace{-2em}
\label{fig:confusion}
\end{wrapfigure}

\topic{\ours{} less relies on co-occurrence prior.}
In \figref{validation} (a), we present the False Seen Prediction (FSP) and False Co-occurrence Prediction (FCP) curves of \ours{} and the baseline~\cite{li2024c2c} on Sth-com. 
While the baseline’s FSP and FCP increase substantially during training (53\% $\rightarrow$ 63\% and 26\% $\rightarrow$ 29\%), those of \ours{} remain consistently lower and even decrease (47\% $\rightarrow$ 44\% and 23\% $\rightarrow$ 23\%), indicating markedly less dependence on co-occurrence statistics.

\topic{Improved unseen generalization.}
In \figref{validation} (b), we present the learning curve of our \ours{} and the baseline~\cite{li2024c2c} on Sth-com. 
\ours{} significantly improves unseen composition accuracy upon the baseline (34\% vs. 30\%), and reduces the seen--unseen accuracy gap (9 points vs. 17 points).

\topic{The learned verb representations are indeed effective.}
Inspired by prior work~\cite{yun2022time}, we evaluate models on the \emph{Temporal} subset of Sth-com~\cite{li2024c2c}.
Specifically, to measure a model’s dependence on static cues, we compare its performance when using temporally shuffled verb features $\f^V_{\text{shuffled}}$ versus the original features $\f^V$.
As shown in \figref{validation} (c), across both seen and unseen compositions, \ours{} exhibits a substantially larger performance gap between original and shuffled features. 
This demonstrates that its verb representations are effectively grounded in temporal dynamics.
In \figref{confusion}, we visualize verb confusion matrices on \emph{unseen} compositions of the Sth-com test set.
Unlike C2C~\cite{li2024c2c}, which frequently confuses opposite verbs (\eg, `unfolding' vs.\ `folding'), \ours{} discriminates them well, demonstrating that it effectively mitigates object-driven shortcuts in verb learning.

\subsection{Comparisons with State-of-the-Art}
\label{sec:main_results}
\begin{table*}[t]
\centering
\vspace{-1em}
\caption{\textbf{Sth-com~\cite{li2024c2c} results.} 
    We show the top-1 verb, object, and composition classification accuracy (\%). 
    We report performance on both seen and unseen compositions, along with their harmonic mean (H.M.).
}
\vspace{-.5em}
\def\arraystretch{1.2}
\resizebox{\linewidth}{!}{
\begin{tabular}{ll ccc c ccc c ccc}
\toprule
\multirow{3}{*}{Backbone} & \multirow{3}{*}{Method} 
& \multicolumn{3}{c}{Verb} && \multicolumn{3}{c}{Object} && \multicolumn{3}{c}{Composition} \\ 
\cline{3-5} \cline{7-9} \cline{11-13} 
&& \multirow{2}{*}{\shortstack{\rule{0pt}{2.8ex}@Seen\\Comp}} & \multirow{2}{*}{\shortstack{\rule{0pt}{2.8ex}@Unseen\\Comp}} & \multirow{2}{*}{H.M.} && \multirow{2}{*}{\shortstack{\rule{0pt}{2.8ex}@Seen\\Comp}} & \multirow{2}{*}{\shortstack{\rule{0pt}{2.8ex}@Unseen\\Comp}} & \multirow{2}{*}{H.M.} && \multirow{2}{*}{Seen (\cg{}$\uparrow$)} & \multirow{2}{*}{Unseen (\cg{}$\uparrow$)} & \multirow{2}{*}{H.M.} \\
\\
\midrule
\multirow{4}{*}{CLIP} & AIM~\cite{yang2023aim} & 54.99 & 39.19 & 45.76 && 67.19 & 53.43 & 59.53 && 40.24 (+3.29) & 18.50 ($-$2.44) & 25.35 \\
& C2C~\cite{li2024c2c} & 63.60 & 54.36 & 58.62 && 67.72 & 56.10 & \textbf{61.36} && 46.31 (+3.24) & 30.08 ($-$0.42) & 36.47 \\
& Jung et al.~\cite{jung2025crr} & 66.57 & 53.55 & 59.35 && 66.56 & 50.90 & 57.69 && \textbf{48.41} (+4.10) & 25.25 ($-$2.01) & 33.19 \\
& \ours{} & 65.73 & 59.00 & \textbf{62.18} && 64.79 & 56.34 & 60.27 && 44.99 (+2.40) & \textbf{33.90} (+0.66) & \textbf{38.67} \\

\midrule
\multirow{3}{*}{\shortstack{InternVideo2}} & LoRA~\cite{hu2022lora} & 48.83 & 36.50 & 41.78 && 64.79 & 58.61 & 61.55 && 32.83 (+1.19) & 19.71 ($-$1.68) & 24.63 \\
& C2C~\cite{li2024c2c} & 70.35 & 63.29 & 66.63 && 70.03 & 63.44 &\textbf{66.58} && \textbf{51.50} (+2.24) & 39.53 ($-$0.63) & 44.73 \\
& \ours{} & 71.65 & 66.65 & \textbf{69.06} && 67.96 & 64.56 & 66.22 && 50.20 (+1.51) & \textbf{43.98} (+0.95) & \textbf{46.88} \\
\bottomrule
\end{tabular}
}

\label{tab:sth_com_results}
\end{table*}

\topic{Sth-com.} As shown in \tabref{sth_com_results}, \ours{} achieves clear gains on Sth-com~\cite{li2024c2c} under a rigorous open-world setting. 
Compared to the C2C~\cite{li2024c2c} baseline, \ours{} improves the verb@unseen-comp by 4.6 points and the unseen composition accuracy by 3.8 points with the CLIP~\cite{radford2021clip} backbone, and by 3.4 points and 4.5 points with InternVideo2 (IV2)~\cite{wang2024iv2} backbone, respectively.
These gains are consistent with stronger verb recognition observed in our analysis (\secref{validation}).
\vspace{-1.2em}
\begin{wraptable}{r}{0.3\textwidth}
\centering
\vspace{-2em}
\caption{\textbf{Closed-world results on Sth-com\cite{li2024c2c}.} 
}
\resizebox{\linewidth}{!}{
\begin{tabular}{l c cc}
\toprule
\multirow{2}{*}{Method} 
&& \multicolumn{2}{c}{Closed-world} \\
\cline{3-4}
&& H.M. & AUC \\
\midrule
C2C~\cite{li2024c2c} && 42.8 & 23.8 \\
Jung et al.~\cite{jung2025crr} && 42.6 & 23.5 \\
\ours{} && \textbf{43.2} & \textbf{25.3} \\
\midrule
C2C~(enhance)~\cite{li2024c2c} && 44.5 & 26.0 \\
LogicCAR~\cite{ye2025logic} && 45.2 & 27.0 \\
\ours{}~(enhance) && \textbf{46.1} & \textbf{27.5} \\
\bottomrule
\end{tabular}
}
\vspace{-2.5em}
\label{tab:sth_com_close}
\end{wraptable}

\noindent Notably, \ours{} yields a positive compositional gap (\cg{}) on unseen compositions, whereas all baselines remain negative, indicating improved compositional behavior on unseen pairs.
Jung et al.~\cite{jung2025crr} yields the lowest unseen composition accuracy among the CLIP baselines while showing the highest seen composition accuracy. Compared to C2C~\cite{li2024c2c}, it improves seen composition accuracy by 2.1 points but degrades unseen composition accuracy by 4.8 points, reflecting a strong bias toward seen compositions and limited open-world generalization.
In \tabref{sth_com_close}, \ours{} achieves the highest H.M. and AUC within each backbone group under the closed-world, test-set-tuned protocol used in prior work~\cite{li2024c2c, jung2025crr, ye2025logic}.

\topic{EK100-com.}
Despite the strong co-occurrence prior in the EK100-com training compositions, \ours{} achieves higher unseen composition accuracy and the best H.M.
As shown in \tabref{ek100_com_results}, compared to C2C~\cite{li2024c2c}, \ours{} substantially improves unseen composition accuracy---by 6.9 points with the CLIP~\cite{radford2021clip} backbone and 7.0 points with the IV2~\cite{wang2024iv2} backbone---despite a decrease in seen composition accuracy.
This pattern reflects reduced reliance on co-occurrence statistics and improved recognition of unseen compositions.
Moreover, \ours{} yields a positive \cg{} on unseen compositions (+1.32 with CLIP and +3.83 with IV2), which is consistent with improved compositional behavior under highly sparse supervision.
In the case of Jung et al.~\cite{jung2025crr}, the model yields lower composition accuracy than C2C~\cite{li2024c2c} on both seen ($-$3.2 points) and unseen ($-$4.6 points) compositions with the CLIP~\cite{radford2021clip} backbone, suggesting that its benefit may be limited under the highly sparse compositional supervision of EK100-com.

\begin{table*}[t]
\centering
\caption{\textbf{EK100-com results.} 
    We show the top-1 verb, object, and composition classification accuracy (\%). 
    We report performance on both seen and unseen compositions, along with their harmonic mean (H.M.).
}
\vspace{-1em}

\def\arraystretch{1.2}
\resizebox{\linewidth}{!}{
\begin{tabular}{ll ccc c ccc c ccc}
\toprule
\multirow{3}{*}{Backbone} & \multirow{3}{*}{Method} 
& \multicolumn{3}{c}{Verb} && \multicolumn{3}{c}{Object} && \multicolumn{3}{c}{Composition} \\ 
\cline{3-5} \cline{7-9} \cline{11-13} 
&& \multirow{2}{*}{\shortstack{\rule{0pt}{2.8ex}@Seen\\Comp}} & \multirow{2}{*}{\shortstack{\rule{0pt}{2.8ex}@Unseen\\Comp}} & \multirow{2}{*}{H.M.} && \multirow{2}{*}{\shortstack{\rule{0pt}{2.8ex}@Seen\\Comp}} & \multirow{2}{*}{\shortstack{\rule{0pt}{2.8ex}@Unseen\\Comp}} & \multirow{2}{*}{H.M.} && \multirow{2}{*}{Seen (\cg{}$\uparrow$)} & \multirow{2}{*}{Unseen (\cg{}$\uparrow$)} & \multirow{2}{*}{H.M.} \\
\\
\midrule
\multirow{4}{*}{CLIP} &  %
AIM~\cite{yang2023aim} & 59.97 & 39.99 & 47.98 && 56.23 & 46.76 & 51.06 && 38.58 (+4.86) & 14.98 ($-$3.71) & 21.58 \\
& C2C~\cite{li2024c2c} & 65.52 & 49.57 & 56.44 && 56.54 & 46.99 & 51.32 && \textbf{42.70} (+5.65) & 21.56 ($-$1.73) & 28.65 \\
& Jung et al.~\cite{jung2025crr} & 63.38 & 47.63 & 54.39 && 55.85 & 42.12 & 48.02 && 39.50 (+4.10) & 16.96 ($-$3.10) & 23.73 \\
& \ours{} & 66.19 & 54.31 & \textbf{59.66} && 54.35 & 49.88 & \textbf{52.02} && 39.70 (+3.73) & \textbf{28.41} (+1.32) & \textbf{33.12} \\
\midrule
\multirow{3}{*}{InternVideo2} & LoRA~\cite{hu2022lora} & 59.73 & 32.99 & 42.51 && 51.08 & 39.34 & 44.45 && 36.43 (+5.92) & 5.63 ($-$7.35) & 9.75\\
& C2C~\cite{li2024c2c} & 68.95 & 57.01 & 62.41 && 60.54 & 52.31 & 56.12 && \textbf{47.30} (+5.55) & 30.33 (+0.51) & 36.96 \\
& \ours{} & 67.76 & 59.78 & \textbf{63.52} && 56.28 & 56.00 & \textbf{56.14} && 41.03 (+2.89) & \textbf{37.31} (+3.83) & \textbf{39.08} \\
\bottomrule
\end{tabular}
}
\label{tab:ek100_com_results}
\end{table*}

\subsection{Ablation Studies}
\label{sec:ablation}

\topic{Effects of \ours{} components.}
In \tabref{ablation_ours} (a),
using \torc{} alone yields the largest gain in verb generalization, improving \textit{verb@unseen-comp} by 3.5 points, and gain in unseen composition accuracy by 3.8 points, compared to the baseline without CPR and TORC (C2C~\cite{li2024c2c}).
Using \cpr{} alone primarily improves unseen composition accuracy by 3.1 points, while reducing seen composition accuracy 4.2 points, indicating a seen--unseen trade-off.
Combining \cpr{} and \torc{} achieves the best overall results, improving unseen composition accuracy by +5.0 points and composition H.M. by +2.7 points, suggesting complementary benefits of mitigating co-occurrence bias and enforcing temporal sensitivity.

\topic{Effects of \torc{} components.}
\tabref{ablation_torc} (b) evaluates the roles of $L_{\cos}$, which enforces discrimination of opposite temporal semantics, and $L_\text{ent}$, which prevents reliance on static cues.
Each term individually improves performance, while using both terms yields the largest gains in the verb@unseen-comp (+3.5 points) and unseen composition accuracy (+3.8 points).

\begin{table}[t]
\centering

\caption{\textbf{Ablation study on the Sth-com~\cite{li2024c2c} validation set.} 
    We use C2C~\cite{li2024c2c} with the CLIP~\cite{radford2021clip} backbone as our baseline, using the unbiased open-world setting.
    We report performance on both seen and unseen compositions, along with their harmonic mean (H.M.).
    V@S and V@U denotes verb@seen-comp and verb@unseen-comp, while O@S and O@U denotes object@seen-comp and object@unseen-comp accuracies.
}
\vspace{-1em}
\mpage{0.48}{
{%
(a) Effects of \ours{} components.}

\resizebox{\linewidth}{!}{
\begin{tabular}{cc cc c cc c ccc}
\toprule
\multirow{2}{*}{\cpr{}} & \multirow{2}{*}{\torc{}} 
& \multicolumn{2}{c}{Verb} && \multicolumn{2}{c}{Object} && \multicolumn{3}{c}{Composition} \\ 
\cline{3-4} \cline{6-7} \cline{9-11} 
&& V@S & V@U && O@S & O@U && Seen & Unseen & H.M. \\
\midrule
&& 65.26 & 54.02 && 67.21 & 56.81 && 47.31 & 30.14 & 36.83 \\
\checkmark & & 63.95 & 54.05 && 63.97 & 59.52 && 43.09 & 33.22 & 37.51 \\
& \checkmark & 64.93 & 57.49 && 66.20 & 57.23 && 46.28 & 33.92 & 39.15 \\
\checkmark & \checkmark & 67.06 & 58.13 && 63.67 & 58.66 && 45.00 & 35.16 & \textbf{39.48} \\
\bottomrule
\end{tabular}
\label{tab:ablation_ours}
}
}
\mpage{0.48}{
{
(b) Effects of \torc{} components.
}
\resizebox{\linewidth}{!}{
\begin{tabular}{cc cc c cc c ccc}
\toprule
\multirow{2}{*}{$L_{\cos}$} & \multirow{2}{*}{$L_\text{ent}$} 
& \multicolumn{2}{c}{Verb} && \multicolumn{2}{c}{Object} && \multicolumn{3}{c}{Composition (OW)} \\ 
\cline{3-4} \cline{6-7} \cline{9-11} 
&& V@S & V@U && O@S & O@U && Seen & Unseen & H.M. \\
\midrule
&& 65.26 & 54.02 && 67.21 & 56.81 && 47.31 & 30.14 & 36.83 \\
\checkmark & & 64.15 & 55.96 && 67.71 & 55.97 && 47.13 & 31.11 & 37.48 \\
& \checkmark & 64.84 & 54.82 && 67.57 & 56.23 && 47.46 & 30.84 & 37.39 \\
\checkmark & \checkmark & 64.93 & 57.49 && 66.20 & 57.23 && 46.28 & 33.92 & \textbf{39.15} \\
\bottomrule
\end{tabular}
\label{tab:ablation_torc}
}
}
\mpage{0.48}{
{
(c) Effects of label space in \cpr{}.
}
\resizebox{\linewidth}{!}{
\begin{tabular}{l ccc ccc ccc}
\toprule
\multirow{2}{*}{Label Space} & \multicolumn{2}{c}{Verb} && \multicolumn{2}{c}{Object} && \multicolumn{3}{c}{Composition} \\
\cline{2-3} \cline{5-6} \cline{8-10}
& V@S & V@U && O@S & O@U && Seen & Unseen & H.M. \\
\midrule
Closed-world ($\mathbb{Y}_\text{seen}$) &  64.61 & 54.26 && 67.21 & 57.46 && 46.78 & 30.80 & 37.14 \\
Full Open-world ($\mathbb{Y}^V \times \mathbb{Y}^O$) & 67.36 & 45.01 && 68.49 & 52.29 && 50.72 & 17.34 & 25.85 \\
Batch-adaptive ($\mathbb{Y}_\text{exp}$) & 64.17 & 54.55 && 63.60 & 58.70 && 42.74 & 33.05 & \textbf{37.28} \\
\bottomrule
\end{tabular}
\label{tab:ablation_label_cpr}
}
}
\mpage{0.48}{
{
(d) Effects of penalty set in \cpr{}.
}
\resizebox{\linewidth}{!}{
\begin{tabular}{l ccc ccc ccc}
\toprule
\multirow{2}{*}{Penalty Set $\mathbb{Y}_{\text{pen}}$ Choice } & \multicolumn{2}{c}{Verb} && \multicolumn{2}{c}{Object} && \multicolumn{3}{c}{Composition}  \\
\cline{2-3} \cline{5-6} \cline{8-10}
& V@S & V@U && O@S & O@U && Seen & Unseen & H.M. \\
\midrule
None (w/o $L_{\text{CPR}}$) & 64.17 & 54.55 && 63.60 & 58.70 && 42.74 & 33.05 & 37.28 \\
All Hard Negatives ($\mathbb{Y}_\text{hard}$) & 61.95 & 57.22 && 47.99 & 49.05 && 31.85 & 30.82 & 31.33 \\
Frequent Hard Negatives ($\mathbb{Y}_\text{hard} \cap \mathbb{Y}_\text{freq}^\text{CPR}$) & 65.36 & 55.86 && 63.20 & 57.62 && 43.60 & 33.16 & \textbf{37.67} \\
\bottomrule
\end{tabular}
\label{tab:ablation_penalty_cpr}
}
}
\label{tab:ablation}
\end{table}

\topic{Effects of label space in \cpr{}.}
In \tabref{ablation_label_cpr} (c), we validate the batch-adaptive label space expansion in \cpr{}.
Compared to closed-world training over $\mathbb{Y}_{\text{seen}}$, our batch-adaptive label space improves unseen composition accuracy (+2.3 points) and H.M. by injecting supervision for synthesized compositions while avoiding unstable optimization.
In contrast, naively optimizing cross-entropy over the full $\mathbb{Y}^V \times \mathbb{Y}^O$ space severely hurts unseen composition accuracy by treating most unseen compositions as negatives throughout training.

\topic{Effects of penalty set in \cpr{}.}
In \tabref{ablation_penalty_cpr} (d), we study the effect of the penalty set $\mathbb{Y}_{\text{pen}}$.
Compared to disabling $L_{\text{CPR}}$, penalizing \emph{frequent} hard negatives ($\mathbb{Y}_{\text{hard}}\cap\mathbb{Y}_{\text{freq}}^\text{CPR}$) yields the best trade-off, improving unseen composition accuracy and achieving the highest H.M. of 37.7\%.
In contrast, penalizing all hard negatives degrades object prediction and hurts overall composition performance, suggesting that restricting penalties to frequent confounders is critical.

\section{Conclusions}
\label{sec:conclusions}

In this work, we revisit zero-shot compositional action recognition and identify \emph{object-driven shortcuts} as a key failure mode.
We attribute this behavior to compositional sparsity and asymmetric verb–object learning difficulty.
Using our diagnostic metrics, we show that existing models overfit to co-occurrence priors, which degrades verb learning and thereby harms generalization to unseen compositions.
We propose \ours{} comprising two components: \cpr{}, which expands supervision to synthesized compositions,
and \torc{}, which enforces temporal-order sensitivity.
Across two benchmarks under an open-world evaluation protocol without test-tuned bias calibration, \ours{} consistently improves performance on unseen compositions, 
moving ZS-CAR toward more reliable verb--object compositional reasoning.

\section*{Acknowledgements}
This work was supported by the NAVER Cloud Corporation and partly supported by the Institute of Information \& Communications Technology Planning \& Evaluation(IITP) grant funded by the Korea Government(MSIT) under grant RS-2024-00353131 (25\%) and Information \& Communications Technology Planning \& Evaluation(IITP)-ITRC(Information Technology Research Center) grant funded by the Korea government(MSIT) (IITP-2025-RS-2023-00259004, 25\%).
Additionally, it was supported by the National Research Foundation of Korea (NRF) grant funded by the Korea government(MSIT) (RS-2025-02216217, 50\%).

\bibliographystyle{splncs04}
\bibliography{main}

\clearpage


\appendix
\renewcommand{\theHsection}{appendix.\Alph{section}}
\phantomsection
\section*{\Large \bf Appendix}

In this appendix, we provide comprehensive evaluation-protocol/
dataset/implementation details and additional results to complement the main paper. 
We organize the appendix as follows:

\begin{enumerate}
    \item Overview of our evaluation setting (\secref{zscar_setting}).
    \item Details on EK100-com dataset (\secref{ek100-com}).
    \item Complete implementation details (\secref{imple_detail}).
    \item Additional evidence of object-driven shortcuts (\secref{additional_evidence}).
    \item Additional results (\secref{additional_results}).
\end{enumerate}

\section{Overview of our evaluation setting}
\label{sec:zscar_setting}
\begin{figure}[t!]
\centering
\includegraphics[width=.7\linewidth]{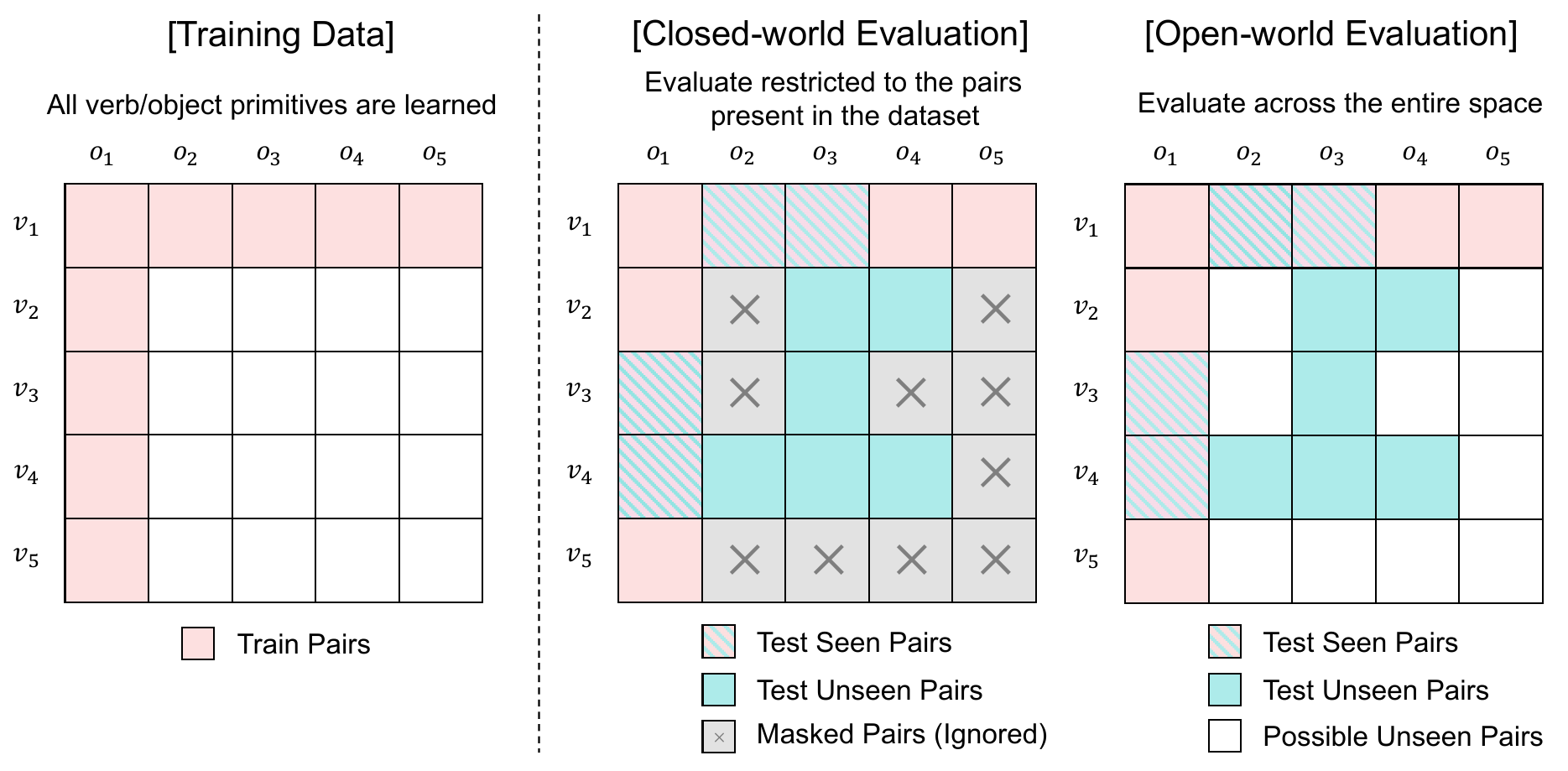} \\
{(a) Closed-world and open-world evaluation settings} \\
\vspace{.8em}
\includegraphics[width=.8\linewidth]{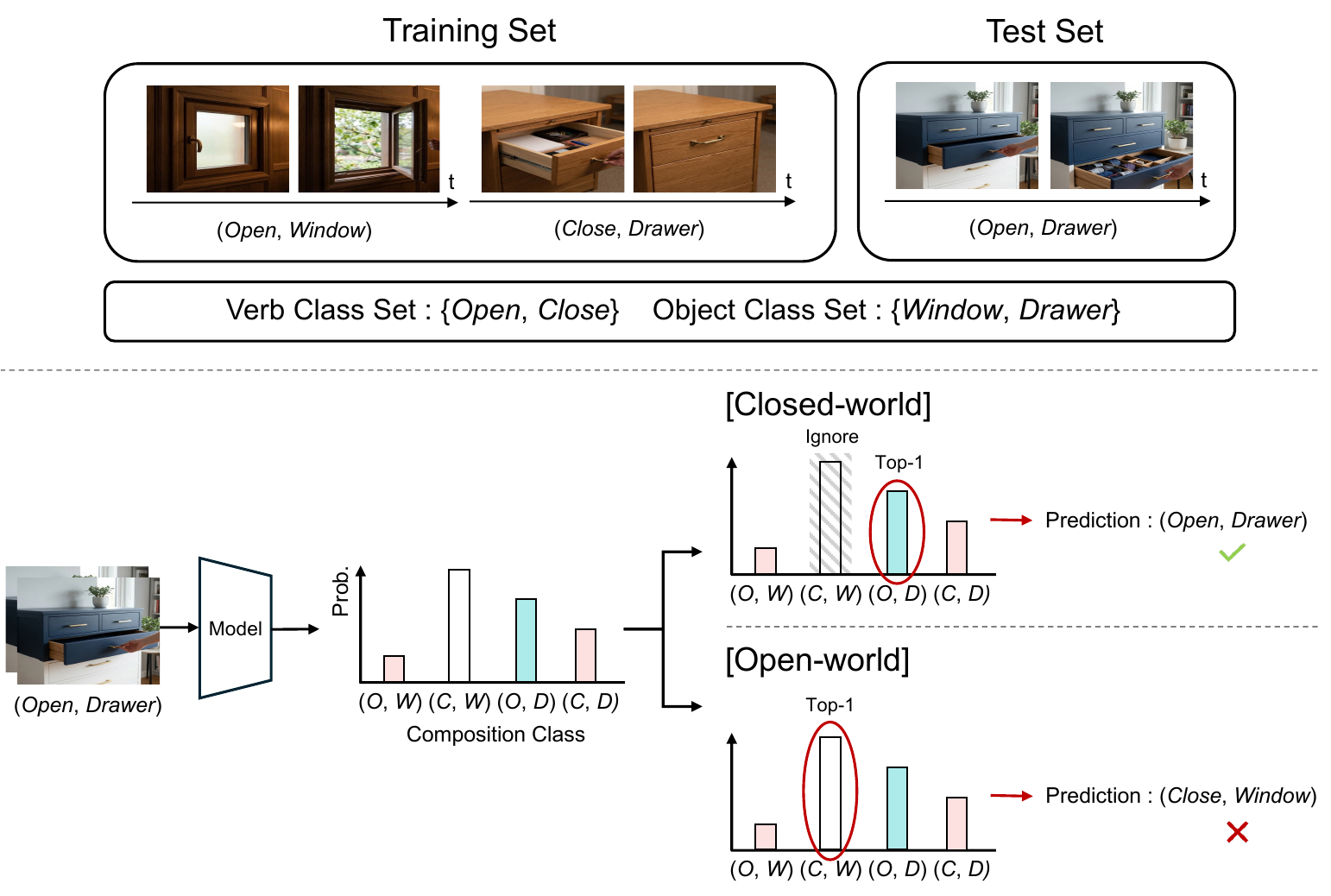} \\
{(b) An example of closed/open-world evaluation} \\
\figcaption{Closed-world and open-world evaluation protocols in ZS-CAR}{
(a) (Left) Primitives are learned from a subset of training compositions (\textit{Train Pairs}). 
(Middle) Closed-world evaluation restricts inference to dataset-present pairs (\textit{Train Pairs} and \textit{Test Seen/Unseen Pairs}), explicitly ignoring others as \textit{Masked Pairs}. 
(Right) Open-world evaluation spans the entire compositional space, requiring the model to navigate all \textit{Possible Unseen Pairs} without artificial masking.
(b) In the closed-world setting, the \textit{Possible Unseen Pair}---not appearing in the dataset---(\textit{Close}, \textit{Window}) becomes a \textit{Masked Pair}, and its logit is ignored.
Consequently, the next valid class (\textit{Open}, \textit{Drawer}) is selected as the top-1 prediction. 
This closed-world setting constrains the evaluation of numerous possible-but-unseen compositions. However, in the open-world setting, the model treats all pairs as valid, thus predicting (\textit{Close}, \textit{Window}) as the top-1 prediction.
}
\label{fig:setting}
\end{figure}

\begin{figure}[h!]
\centering
\includegraphics[width=\linewidth]{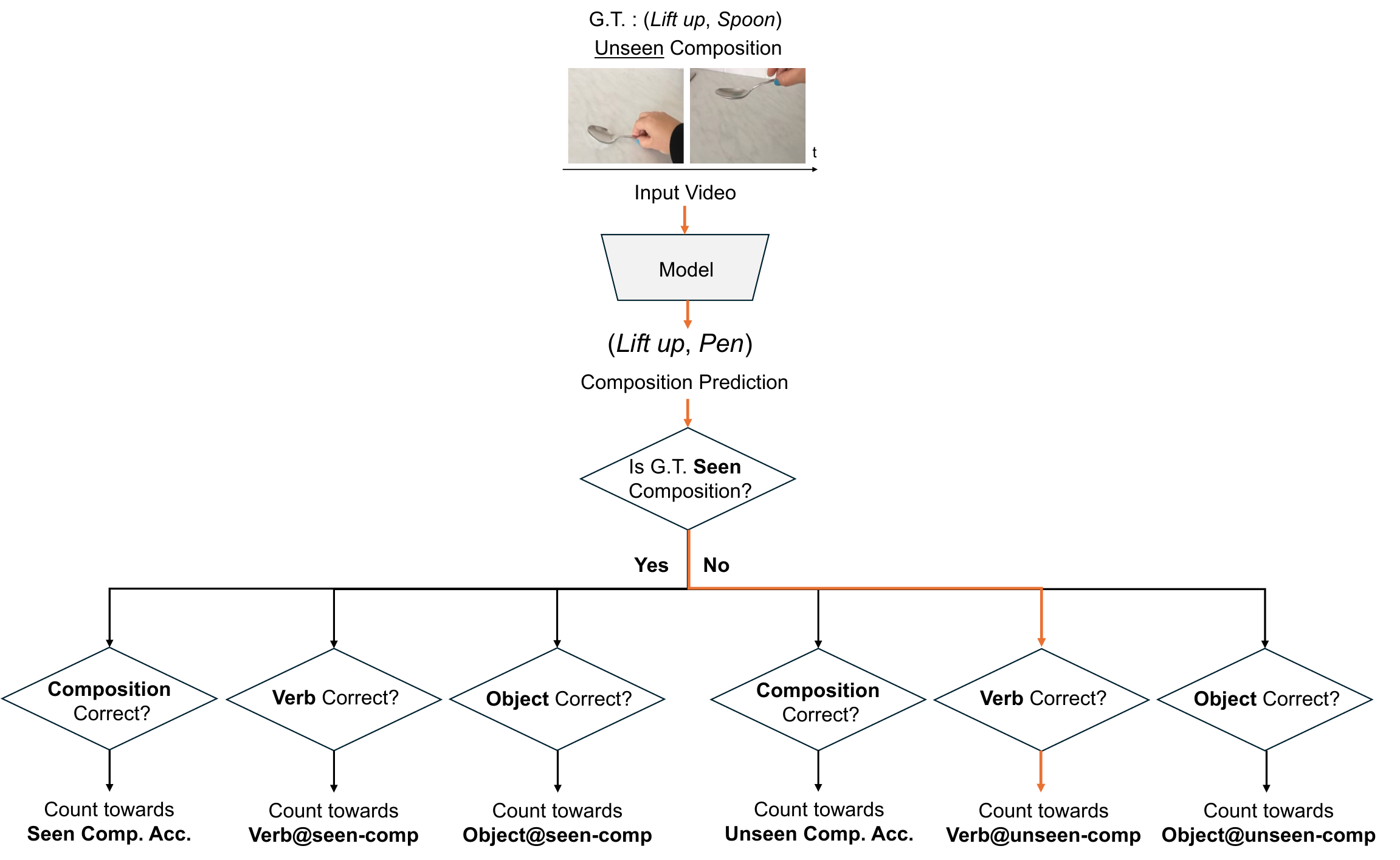} \\
\figcaption{Evaluation metrics used in this work}{
We evaluate accuracies for verbs, objects, and compositions on seen and unseen composition splits separately. 
Based on the model's composition predictions, we count the correct matches at each level to report (1) seen/unseen composition accuracy, (2) verb@seen/unseen-comp, and (3) object@seen/unseen-comp.
For instance, predicting an unseen composition (\textit{Lift up}, \textit{Spoon}) as (\textit{Lift up}, \textit{Pen}) updates only the correct count for verb@unseen-comp (orange arrow).
}
\label{fig:setting_metrics}
\end{figure}

In \figref{setting} and \figref{setting_metrics}, we briefly describe our evaluation setting.
In Zero-Shot Compositional Action Recognition (ZS-CAR), the compositional label space spans the Cartesian product of the verb and object primitives ($|\mathbb{Y}^V| \times |\mathbb{Y}^O|$). 
At test time, samples are categorized into \textit{seen} and \textit{unseen} compositions depending on whether they were observed during training, as shown in \figref{setting} (a). 
Under the \textit{closed-world} evaluation setting, the evaluation is restricted to compositions present within the dataset (i.e., the union of the training and test sets). 
Consequently, the model computes compositional logits exclusively for these predefined closed-world pairs, as shown in \figref{setting} (b).

However, this \textit{closed-world} setting suffers from the impractical assumption of requiring prior knowledge of the test set's ground-truth labels. 
Therefore, we adopt the \textit{open-world} evaluation setting as our primary protocol. 
By evaluating across the entire compositional space and computing logits for all possible combinations, as shown in \figref{setting} (b), the open-world setting enables the evaluation of various possible-but-unseen compositions encountered in real-world applications, where the test distribution is unknown.

As shown in \figref{setting_metrics}, we measure verb, object, and composition accuracies across seen and unseen splits to evaluate not only the model's compositional recognition capability, but also how primitive recognition degrades on unseen compositions. 
Specifically, based on the model's composition predictions, we verify whether the composition, verb, or object is predicted correctly, yielding three sets of metrics: (1) seen/unseen composition accuracy, (2) verb@seen/unseen-comp, and (3) object@seen/unseen-comp.

\section{Details on EK100-com dataset}
\label{sec:ek100-com}

\begin{figure*}[t]
\centering 
\mpage{0.48}{
\includegraphics[width=\linewidth]{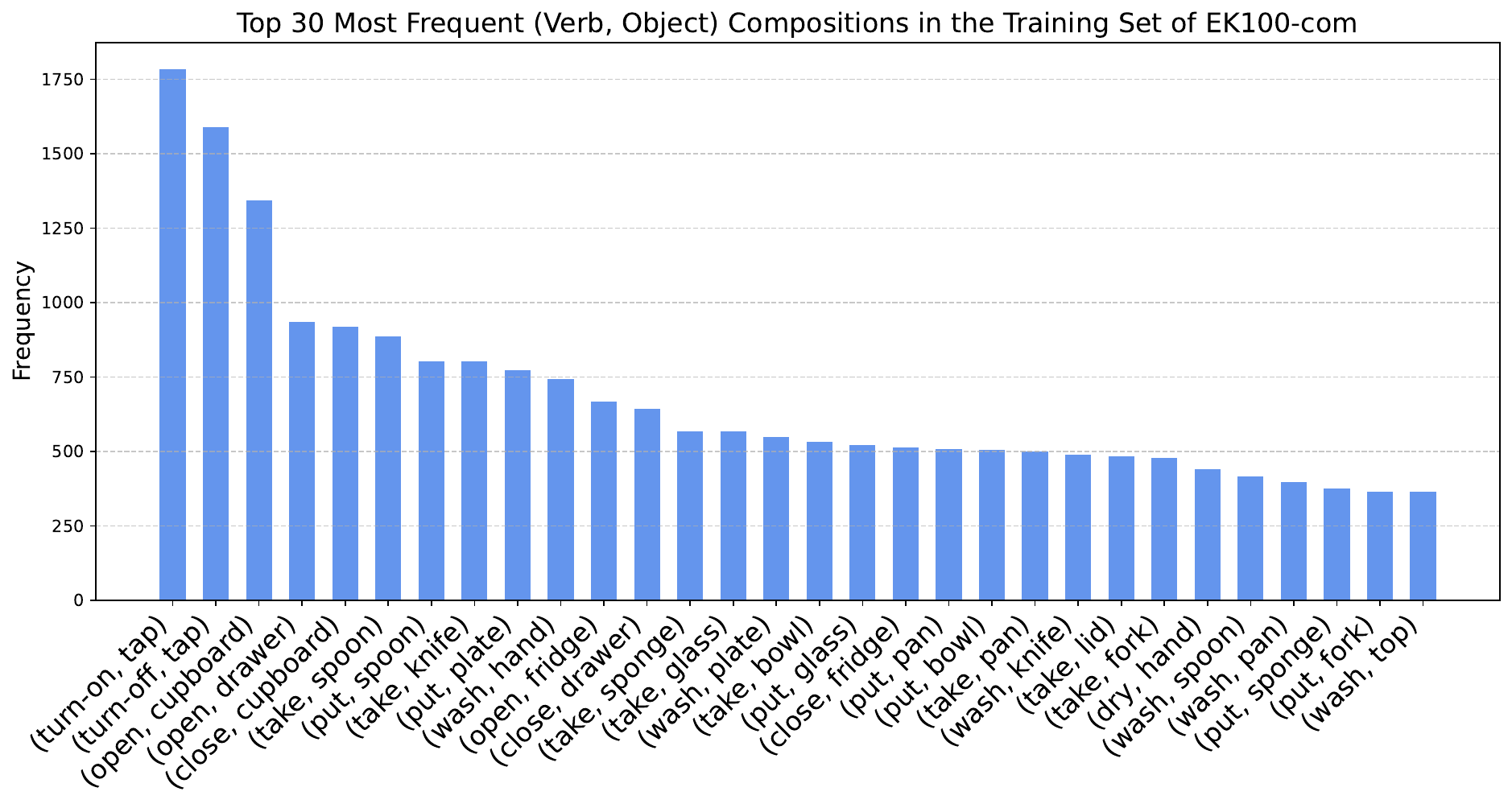}
\\
\vspace{.3em}}
\mpage{0.48}{
\includegraphics[width=\linewidth]{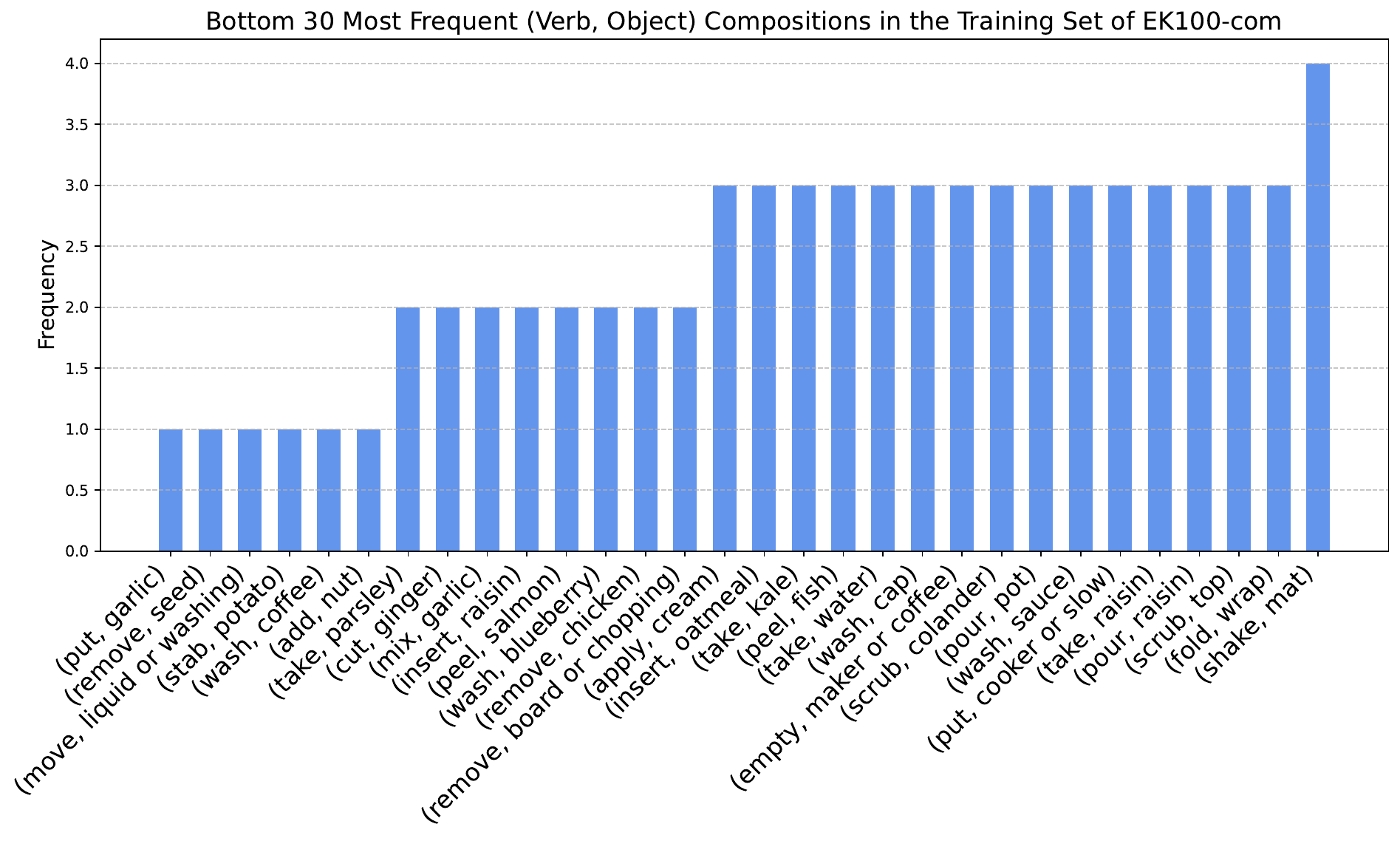}}
\\
\mpage{0.48}{\hspace{.5cm}
Top-30 frequent compositions
}
\mpage{0.48}{\hspace{.5cm}
Bottom-30 frequent compositions
}
\\
\figcapmargin
\figcaption{Top/Bottom-30 frequent compositions in the EK100-com training set}{
}
\label{fig:ekcom_frequent_pairs}
\end{figure*}

\begin{figure}[h]
\centering    
\mpage{0.45}{
\centering    
\includegraphics[width=\linewidth]{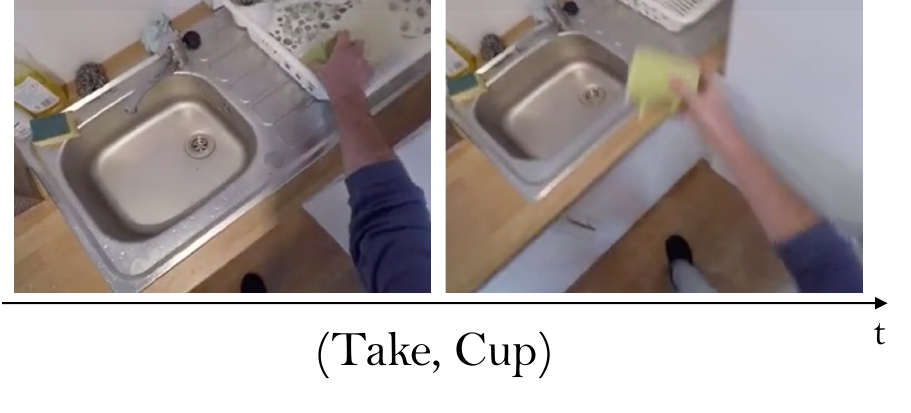}
} 
\mpage{0.45}{
\centering    
\includegraphics[width=\linewidth]{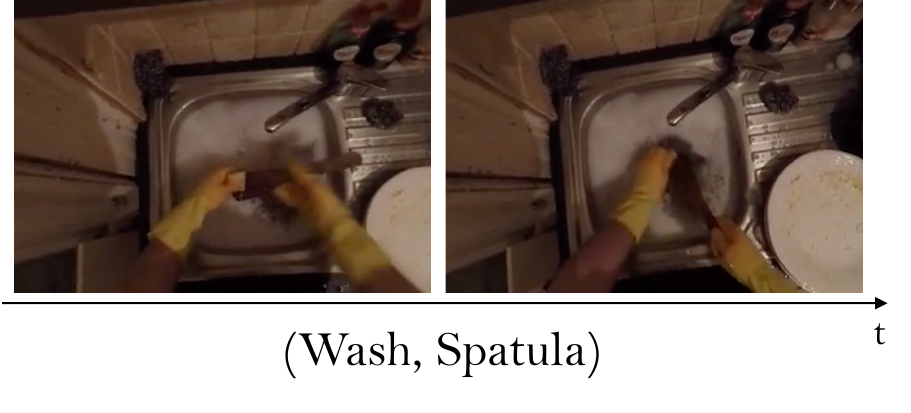}
} 
\\
\vspace{-1em}
{\hspace{.2cm}Seen composition examples}
\\
\vspace{.5em}
\mpage{0.45}{
\centering    
\includegraphics[width=\linewidth]{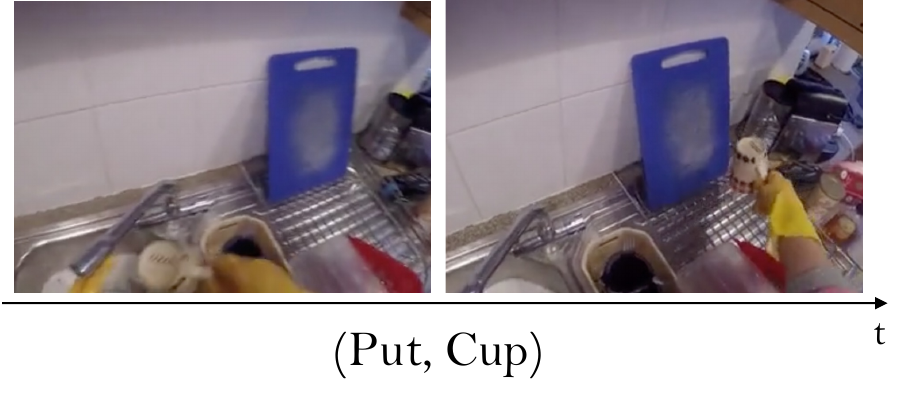}
} 
\mpage{0.45}{
\centering    
\includegraphics[width=\linewidth]{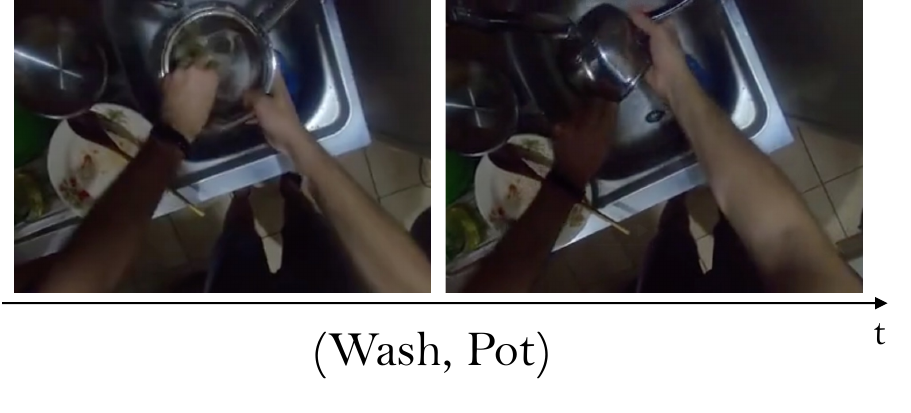}
}
\\
\vspace{-1em}
\hspace{.2cm}{Unseen composition examples}
\\
\figcapmargin
\figcaption{Example of seen/unseen composition samples in the EK100-com dataset}{
}
\label{fig:ekcom_example}
\end{figure}

In this section, we provide details about our curated ZS-CAR benchmark, EPIC-KITCHENS-100-composition (EK100-com).
We construct EK100-com by repurposing EPIC-KITCHENS-100 (EK100)~\cite{damen2022ek100} following the same protocol of constructing Sth-com~\cite{li2024c2c}.
In particular, we use the original training (67,217 samples) and validation (9,668 samples) split of EK100~\cite{damen2022ek100} and split them as follows: 
(1) we filter the data to include only compositions with more than five samples each.
(2) we ensure that all compositions present in the validation set also exist in the training set.
After initial filtering, the training set comprises 62,790 samples with 1,331 unique compositions and the validation set contains 8,657 samples, which consist of 833 seen compositions and 3 unseen compositions.
(3) we randomly select half of the compositions from both the training and validation sets and swap these subsets.
(4) Finally, we split the validation and test set with a ratio of $3:4$.
We provide the statistics of EK100-com in \tabref{ek100-com_setting} and \figref{ekcom_frequent_pairs}, and some seen/unseen composition examples of EK100-com in \figref{ekcom_example}.

Beyond expanding the pool of publicly available ZS-CAR benchmarks, our repurposed EK100-com exhibits two key properties that make it a compelling ZS-CAR benchmark.
First, unlike Something-Something V2~\cite{goyal2017something}, where videos may involve multiple objects associated with a verb, videos in EK100~\cite{damen2022ek100} depict actions consisting of a single verb and a single object. 
Therefore, the training data of EK100-com provides explicit object supervision for the models.
Second, since EK100~\cite{damen2022ek100} has long-tailed distribution, the label coverage ratio---defined as the number of unique verb--object pairs present in the dataset relative to all possible combinations—--of EK100-com is 7.5\%, which is much lower than that of Sth-com~\cite{li2024c2c} (12.8\%).
This makes EK100-com a challenging benchmark, suitable for evaluating whether a model can overcome the risk of shortcut learning induced by data sparsity.

\begin{table}[t]
\centering
    \caption{\textbf{Details of the our introduced EK100-com benchmark.}
    The label coverage ratio of EK100-com is $7.5\%=1320/(81\times216)$.
    }
\resizebox{.7\linewidth}{!}{
\begin{tabular}{l cc ccc}
\toprule
EK100-com & \# Verb & \# Object & \# Compositions & \# Samples \\
\midrule
Train & 81 & 216 & 1133 & 54691 \\
Val & 57 & 175 & 462 Seen \& 187 Unseen & 3085 Seen \& 4007 Unseen  \\
Test & 61 & 177 & 522 Seen \& 187 Unseen & 4197 Seen \& 5258 Unseen \\
\midrule
Total & 81 & 216 & 1320 & 71238 \\
\bottomrule
\end{tabular}
}
\label{tab:ek100-com_setting}
\end{table}

\section{Complete Implementation Details}

\label{sec:imple_detail}

\subsection{Implementation details on our diagnosis}
In this section, we provide comprehensive details of experimental setup and implementation details of our diagnosis for ZS-CAR in Section 3 in the main paper.
All notations used in this appendix are consistent with those defined in the main paper.
As mentioned in the main paper, we adopt the state-of-the-art model, C2C~\cite{li2024c2c}, as the existing ZS-CAR baseline for diagnosis.

\topic{Details about our diagnosis in Figure 2 (a).}
In Figure 2 (a) in the main paper, we empirically demonstrate the asymmetric learning difficulty between verbs and objects.
We train a randomly initialized ViT~\cite{dosovitskiy202vit} with joint Spatio-Temporal (ST) attention on our selected subset of the Sth-com~\cite{li2024c2c} dataset. 
We aim to select a dense subset that eliminates sparsity by ensuring samples exist for every pair, maximizing the number of samples while maintaining a sufficient number of distinct labels.
As a result, the selected subset is composed of 10 verb classes: `Moving sth and sth away from each other', `Moving sth and sth closer to each other', `Moving sth away from sth', `Moving sth closer to sth', `Moving sth down', `Moving sth up', `Pushing sth from left to right', `Pushing sth from right to left', `Pushing sth so that it falls off the table', and `Pushing sth so that it slightly moves'; and 10 object classes: `Ball', `Battery', `Book', `Bottle', `Coin', `Glass', `Knife', `Paper', `Remote', and `Spoon'.
We construct the training set consisting of 2,555 samples from the original Sth-Com~\cite{li2024c2c} training and validation samples, while the validation set comprises 260 samples from the original Sth-Com~\cite{li2024c2c} test set.
For training, we employ a multi-task learning approach using separate linear classifiers for verbs and objects, taking the ViT's CLS token as input. 

\begin{figure}[t]
\centering
\includegraphics[width=.25\linewidth]{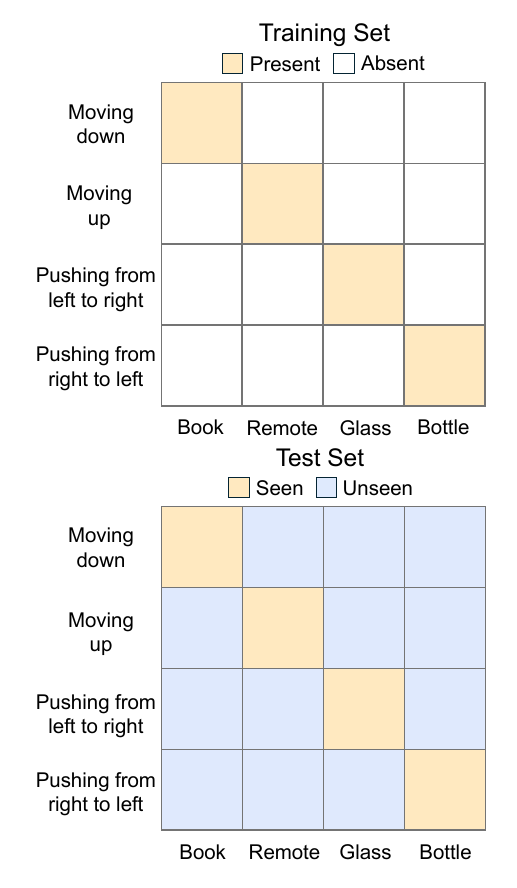}
\raisebox{-.2cm}{\includegraphics[width=.63\linewidth]{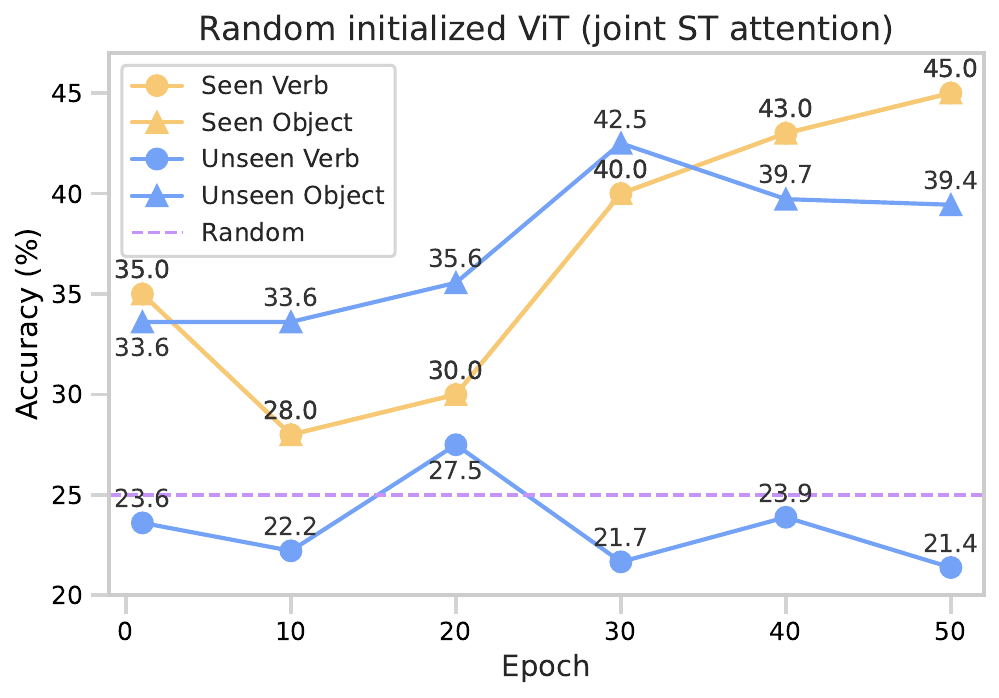}} \\
\vspace{-1em}
\figcapmargin
\figcaption{Object-driven shortcuts with the random initialized model}{
    Even with a randomly initialized model, a ViT-B~\cite{dosovitskiy202vit} with joint ST attention, \textit{object-driven shortcuts emerge}: the model achieves high object accuracy, but verb accuracy on bias-conflicting unseen compositions drops below chance, indicating shortcut-driven verb failures.
}
\label{fig:shortcuts_vit}
\end{figure}

\topic{Details about our diagnosis in Figure 2 (b).}
To precisely control the sample counts for each verb-object pair, we select a $4 \times 4$ toy set from the Sth-Com~\cite{li2024c2c} dataset.
This set contains four verb labels (`Moving sth down', `Moving sth up', `Pushing sth from left to right', and `Pushing sth from right to left') and four object labels (Book, Remote, Glass, Bottle). 
As shown in Figure 2 in the main paper and \figref{shortcuts_vit}, the training set consists of the four diagonal pairs with 40 samples per pair. 
The test set comprises two distinct splits: seen compositions (25 samples per pair) and unseen compositions (30 samples per pair).
We use 8 frames for input videos.
In \figref{shortcuts_vit}, we demonstrate that the object-driven verb shortcut learning observed with CLIP~\cite{radford2021clip} in the main paper emerges identically even in a randomly initialized model. 
This indicates that this phenomenon is not a side effect of pre-trained knowledge, but rather a structural issue inherent to ZS-CAR.

\begin{figure}[t]
\centering  
\mpage{0.48}{
\centering
\includegraphics[width=\linewidth]{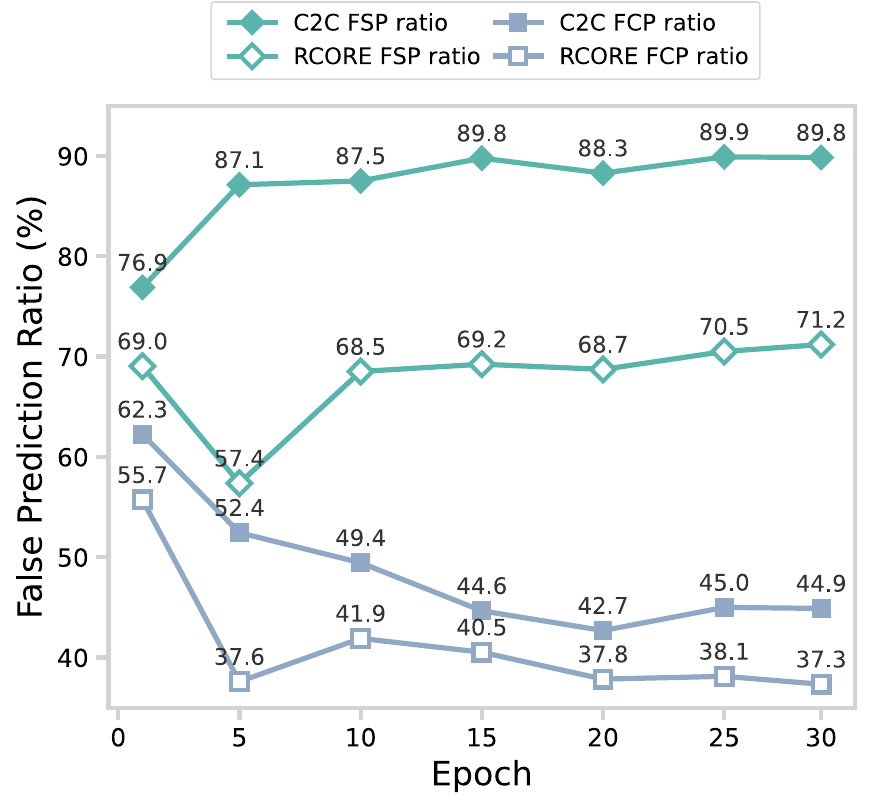}
\\
{\hspace{0.5cm}(a) FSP/FCP}
}
\mpage{0.48}{
\includegraphics[width=\linewidth]{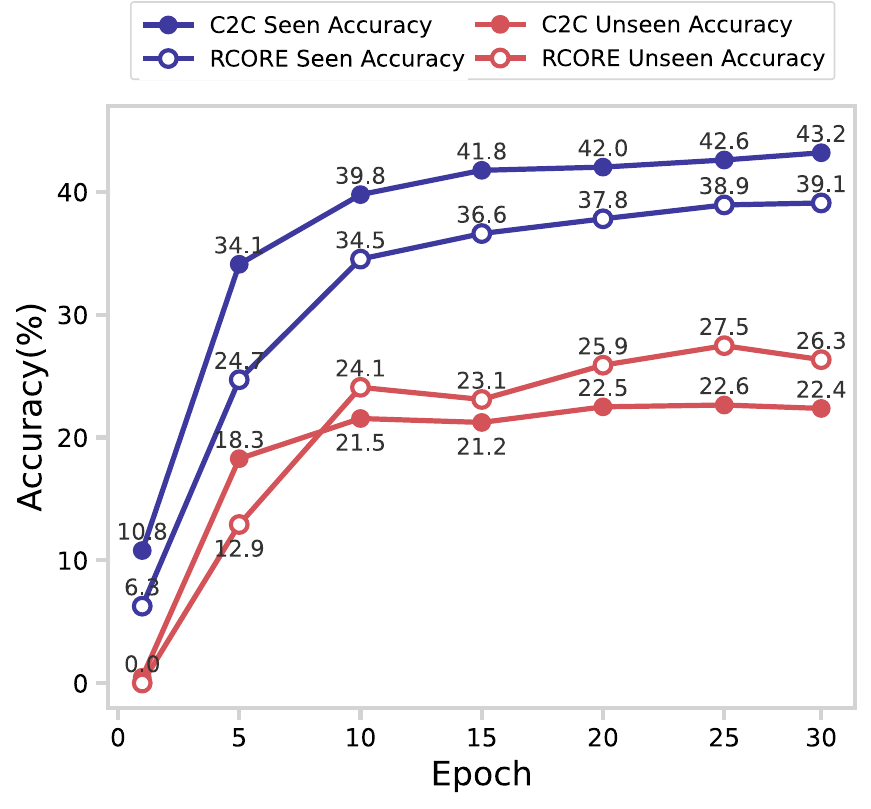}
\\
{\hspace{0.5cm}(b) Composition accuracy}
}
\\
\figcapmargin
\figcaption{Learning curve of the baseline and \ours{} on EK100-com}
{\ours{} (a) suppresses the increase of FCP during training, (b) effectively narrowing the performance gap between seen and unseen composition validation accuracies, compared to the baseline (C2C~\cite{li2024c2c}).
}
\label{fig:ekcom_observation}
\end{figure}

\topic{Details about our diagnosis in Figure 3.}
In Figure 3 in the main paper, we present the learning curves of the baseline model~\cite{li2024c2c} trained on the Sth-com~\cite{li2024c2c}.
We further present diagnostic results for both the baseline and \ours{} on EK100-com in \figref{ekcom_observation}.
We define two misclassification ratios, False Seen Prediction (FSP) and False Co-occurrence Prediction (FCP).
We first collect the misclassified unseen composition samples into the set
\begin{align}
\mathcal{E}_{\mathrm{unseen}}
&= \{i \mid \y_i^C\in\mathbb{Y}_{\mathrm{unseen}},\ \hat{\y}_i^C\neq \y_i^C\}.
\end{align}
FSP measures the proportion of unseen composition validation samples incorrectly classified as seen compositions:
\begin{align}
\vspace{-3em}
\textit{False Seen Prediction}=\frac{1}{|\mathcal{E}_{\mathrm{unseen}}|}
{\sum_{i\in\mathcal{E}_{\mathrm{unseen}}}
\mathbb{I}(\hat{\y}^C_i \in \mathbb{Y}_{\text{seen}})},
\vspace{-.5em}
\end{align}
where $\mathbb{I}(\cdot)$ is the indicator function.
FCP further quantifies, over the same set of misclassified unseen-composition samples, the fraction whose predictions fall into the frequently co-occurring verb–object pairs $\mathbb{Y}_\text{freq}$ in the training data:
\begin{align}
\vspace{-1em}
\textit{False Co-oc. Prediction}=\frac{1}{|\mathcal{E}_{\mathrm{unseen}}|}{\sum_{i\in\mathcal{E}_{\mathrm{unseen}}} \mathbb{I}(\hat{\y}^C_i \in \mathbb{Y}_{\text{freq}})}.
\vspace{-1em}
\end{align}
We further break down these misclassification ratios into the three cases: (i) Verb-collapse, where the predicted object is correct but the verb is incorrect; (ii) Object-collapse, where the predicted verb is correct but the object is incorrect; and (iii) Dual-collapse, where both verb and object are incorrect.
We define them in FSP as follows:
\begin{align}
\textit{Verb-collapse}&=\frac{\sum_{i \in \mathcal{E}_{\mathrm{unseen}}} \mathbb{I}(\hat{\y}^C_i \in \mathbb{Y}_{\text{seen}}) \cdot \mathbb{I}(\hat{\y}^V_i \neq \y^V_i \land \hat{\y}^O_i = \y^O_i)}{\sum_{i \in \mathcal{E}_{\mathrm{unseen}}} \mathbb{I}(\hat{\y}^C_i \in \mathbb{Y}_{\text{seen}})}\\
\textit{Object-collapse}&=\frac{\sum_{i \in \mathcal{E}_{\mathrm{unseen}}} \mathbb{I}(\hat{\y}^C_i \in \mathbb{Y}_{\text{seen}}) \cdot \mathbb{I}(\hat{\y}^V_i = \y^V_i \land \hat{\y}^O_i \neq \y^O_i)}{\sum_{i \in \mathcal{E}_{\mathrm{unseen}}} \mathbb{I}(\hat{\y}^C_i \in \mathbb{Y}_{\text{seen}})}\\
\textit{Dual-collapse}&=\frac{\sum_{i \in \mathcal{E}_{\mathrm{unseen}}} \mathbb{I}(\hat{\y}^C_i \in \mathbb{Y}_{\text{seen}}) \cdot \mathbb{I}(\hat{\y}^V_i \neq \y^V_i \land \hat{\y}^O_i \neq \y^O_i)}{\sum_{i \in \mathcal{E}_{\mathrm{unseen}}} \mathbb{I}(\hat{\y}^C_i \in \mathbb{Y}_{\text{seen}})},
\end{align}
where $\y^C_i=(\y^V_i,\y^O_i)$.
We apply the identical formulation for FCP as well.

\subsection{Implementation details on \ours{}}
\label{sec:imple_detail_ours}

\begin{table}[t]
\centering
\caption{\textbf{Hyperparameters used for training \ours{} on each dataset.} We denote InternVideo2~\cite{wang2024iv2} as `IV2'.
    }
\resizebox{.7\linewidth}{!}{
\begin{tabular}{l cc c cc}
\toprule
\multirow{2}{*}{Config} & \multicolumn{2}{c}{Sth-com~\cite{li2024c2c}} && \multicolumn{2}{c}{EK100-com} \\
\cline{2-3} \cline{5-6}
& CLIP~\cite{radford2021clip} & IV2~\cite{wang2024iv2} && CLIP~\cite{radford2021clip} & IV2~\cite{wang2024iv2} \\
\midrule
Optimizer &  \multicolumn{5}{c}{AdamW~\cite{adamw}} \\
Base learning rate (visual) & 5e-4 & 5e-5 && 1e-4 & 5e-5 \\
Base learning rate (text) & 1e-4 & 5e-5 && 1e-5 & 5e-5 \\
Weight decay (visual) & \multicolumn{5}{c}{1e-4}  \\
Weight decay (text) & \multicolumn{5}{c}{1e-5}  \\
Optimizer momentum &   \multicolumn{5}{c}{{$\beta_1, \beta_2 = 0.9, 0.999$}} \\
Per GPU batch size & \multicolumn{5}{c}{16} \\
Warmup epochs & \multicolumn{5}{c}{3} \\
$\alpha,\beta,\gamma,\delta$ & \multicolumn{5}{c}{0.2, 1.0, 1.0, 0.1} \\
$p_\text{CPR}$ & 0.95 & 0.8 && 0.95 & 0.8 \\
(start, end) epoch for scaling $\gamma$ & \multicolumn{5}{c}{(5, 10)} \\
(start, end) epoch for scaling $\delta$ & (15, 15) & (30, 35) && (35, 40) & (30, 35) \\
Scale factor for $\lambda$ in \cpr{} & \multicolumn{5}{c}{0.5} \\
Softmax temperature & \multicolumn{5}{c}{0.07} \\
Training epochs & 40 & 50 && 40 & 40 \\
\bottomrule
\end{tabular}
}
\label{tab:hyperparam}
\end{table}

\topic{Experimental Setup.} 
We implement our method using PyTorch and conduct all experiments on 8 NVIDIA Tesla V100 GPUs.
For video preprocessing, we uniformly sample frames from each video to generate clips of size $3 \times 16 \times 224 \times 224$ (channels $\times$ frames $\times$ height $\times$ width) for the CLIP~\cite{radford2021clip} backbone and $3 \times 8 \times 224 \times 224$ for the InternVideo2~\cite{wang2024iv2} backbone.

\topic{Model Architecture.}
We utilize CLIP-B/16~\cite{radford2021clip} as our visual/text backbone. 
Following prior work~\cite{li2024c2c}, we adopt AIM~\cite{yang2023aim}, a parameter-efficient spatio-temporal adapter for CLIP~\cite{radford2021clip}, as the video encoder.
We also employ InternVideo2-CLIP-B/14~\cite{wang2024iv2} as our modern video-pretrained VLM backbone and utilize LoRA~\cite{hu2022lora} for parameter-efficient fine-tuning.
We use CoOp-style~\cite{zhou2022coop} learnable text prompts for all models.
We also construct the verb encoder with two temporal convolution layers with ReLU activations, while the object encoder employs temporal average pooling followed by a two-layer MLP with ReLU activations.

To extract $T$ frame-level features with the InternVideo2~\cite{wang2024iv2} backbone, we modify its attention pooling mechanism.
Originally, the model concatenates a CLS token with video tokens in $\mathbb{R}^{NT \times D}$, where $N$ is the number of spatial patches and $D$ is the feature dimension, and applies attention pooling to produce a single video-level representation in $\mathbb{R}^{1 \times D}$.
In our approach, we adapt this to operate at the frame level. Specifically, for each frame, we concatenate the CLS token with the corresponding spatial tokens in $\mathbb{R}^{N \times D}$.
This creates $T$ separate feature sets, each in $\mathbb{R}^{(N + 1) \times D}$.
Finally, we apply attention pooling independently to each of these sets, resulting in the frame-level video features in $\mathbb{R}^{T \times D}$.

\topic{Training Strategy.} 
We train all models, including the baselines and \ours{}, following the hyperparameters detailed in \tabref{hyperparam}, using a total batch size of 128.
For \cpr{}, we employ the unsupervised foreground estimation algorithm FAME~\cite{ding2022fame} to identify high-motion regions. 
FAME~\cite{ding2022fame} isolates the moving foreground from background regions based on frame differences and color statistics. 
To generate the mixed samples, we first sample the mixing coefficient from a $\text{Beta}(2.0, 2.0)$ distribution, following prior work~\cite{yun2019cutmix, zhang2018mixup}.
We then scale it by 0.5 to obtain the final blending coefficient $\lambda$.
We apply this \cpr{}'s augmentation to the samples within a batch with a probability of $p_{\text{aug}}$.
For the frequent verb-object pairs to apply the $L_\text{CPR}$ loss term, we select pairs that possess a strong co-occurrence prior from both verb and object perspectives as the frequent pairs $\mathbb{Y}_\text{freq}^\text{CPR}$, which is defined as $\mathbb{Y}_\text{freq}^\text{CPR} = \mathbb{Y}_\text{freq}^{O \mid V} \cap \mathbb{Y}_\text{freq}^{V \mid O}$.
$\mathbb{Y}_\text{freq}^{O \mid V}$ is the set of pairs where the conditional probability of an object given a verb in the training data is greater than $\mu + \sigma$. 
Here, $\mu$ and $\sigma$ represent the mean and standard deviation of the conditional probabilities calculated on the training set. 
The $\mathbb{Y}_\text{freq}^{V \mid O}$ is defined similarly for the conditional probability of a verb given an object.
$(\mu_{O|V} + \sigma_{O|V}, \mu_{V|O} + \sigma_{V|O})$ are calculated as $(0.13, 0.19)$ for the Sth-com~\cite{li2024c2c} dataset and $(0.23, 0.40)$ for EK100-com.

\topic{Loss configuration.} 
For the total loss, we set the weights for the verb and object components to $\alpha = 0.2$ and for the composition loss to $\beta = 1.0$, following prior work~\cite{li2024c2c}. 
As demonstrated in \tabref{supple_additional_ablation} (d), we empirically find it beneficial to linearly warm up the weights for the \torc{} loss: $\gamma$ is increased from 0.0 to 1.0 over 5 epoch to 10 epoch.
We use $\delta$, the loss scale of $L_\text{CPR}$, differently depending on the dataset and backbone, as detailed in \tabref{hyperparam}.

\topic{Inference.}
During inference, we adopt a single-view, single-crop protocol. 
To select the best model during training and for bias calibration, we use the harmonic mean of the accuracies for seen and unseen compositions as the model selection criterion.

\begin{table*}[t]
\centering
\caption{\textbf{Top-10 failure cases of the existing ZS-CAR model on unseen compositions of the Sth-com test set.} We employ C2C~\cite{li2024c2c} trained on Sth-com~\cite{li2024c2c} as baseline for analysis. We use `sth' instead of `something' to shorten. We report the results with unbiased open-world inference scores. 
`Prediction Rate' denotes the proportion of mispredictions classified as the specific composition. 
`Verb/Object-side' indicates the most frequently co-occurring counterpart for each mispredicted component in the training set, along with its co-occurrence ratio.
The top-10 cases listed below account for 9.5\% of the total mispredictions on unseen compositions.
}

\resizebox{\linewidth}{!}{
\begin{tabular}{l l l l l l l}
\toprule
\multirow{2}{*}{Rank} & \multirow{2}{*}{Ground Truth (Verb, Object)} & Error & \multirow{2}{*}{Top-2 Misprediction (seen/unseen)} & Prediction & \multicolumn{2}{l}{Most Frequent Co-occurring Component During Training} \\ 
\cline{6-7}
&& Count / Rate (\%) && Count / Rate (\%) & Verb-side & Object-side \\
\toprule
\multirow{2}{*}{1} & \multirow{2}{*}{(`Pretending to be tearing [sth that is not tearable]', `Paper')} & \multirow{2}{*}{192 / 99.0\%} & (`Tearing [sth] just a little bit', `Paper') (seen) & 116 / 60.4\% & w/ `Paper' (71.6\%) & w/ `Tearing [sth] just a little bit' (26.7\%) \\
&&& (`Folding [sth]', `Paper') (seen) & 47 / 24.5\% & w/ `Paper' (43.9\%) & w/ `Tearing [sth] just a little bit' (26.7\%) \\
\midrule
\multirow{2}{*}{2} & \multirow{2}{*}{(`Unfolding [sth]', `Paper')} & \multirow{2}{*}{140 / 44.7\%} & (`Folding [sth]', `Paper') (seen) & 80 / 57.1\% & w/ `Paper' (43.9\%) & w/ `Tearing [sth] just a little bit' (26.7\%) \\
&&& (`Tearing [sth] into two pieces', `Paper') (unseen) & 12 / 8.6\% & w/ `Envelope' (17.4\%) & w/ `Tearing [sth] just a little bit' (26.7\%) \\
\midrule
\multirow{2}{*}{3} & \multirow{2}{*}{(`Opening [sth]', `Box')} & \multirow{2}{*}{106 / 82.8\%} & (`Showing that [sth] is inside [sth]', `Box') (seen) & 27 / 25.5\% & w/ `Box' (22.7\%) & w/ `Showing that [sth] is empty' (10.1\%) \\
&&& (`Showing that [sth] is empty', `Box') (seen) & 23 / 21.7\% & w/ `Cup' (30.8\%) & w/ `Showing that [sth] is empty' (10.1\%) \\
\midrule
\multirow{2}{*}{4} & \multirow{2}{*}{(`Squeezing [sth]', `Plastic bag')} & \multirow{2}{*}{94 / 40.9\%} & (`Squeezing [sth]', `Bag') (seen) & 77 / 81.9\% & w/ `Sponge' (11.0\%) & w/ `Stuffing [sth] into [sth]' (15.4\%) \\
&&& (`Folding [sth]', `Plastic bag') (unseen) & 13 / 13.8\% & w/ `Paper' (43.9\%) & w/ `[sth] falling like a feather or paper' (57.4\%) \\
\midrule
\multirow{2}{*}{5} & \multirow{2}{*}{(`Putting [sth] into [sth]', `Box')} & \multirow{2}{*}{88 / 60.3\%} & (`Stuffing [sth] into [sth]', `Box') (seen) & 10 / 11.4\% & w/ `Box' (19.2\%) & w/ `Showing that [sth] is empty' (10.1\%) \\
&&& (`Putting [sth] into [sth]', `Plastic box') (seen) & 8 / 9.1\% & w/ `Bowl' (12.4\%) & w/ `Closing [sth]' (25.9\%) \\
\midrule
\multirow{2}{*}{6} & \multirow{2}{*}{(`Lifting up one end of [sth], then letting it drop down', `Card')} & \multirow{2}{*}{79 / 54.1\%} & (`Lifting [sth] up completely without letting it drop down', `Card') (unseen) & 20 / 25.3\% & w/ `Crayon' (26.8\%) & w/ `Uncovering [sth]' (19.8\%) \\
&&& (`Lifting up one end of [sth], then letting it drop down', `Pencil') (unseen) & 19 / 24.1\% & w/ `Crayon' (26.8\%) & w/ `Uncovering [sth]' (13.7\%) \\
\midrule
\multirow{2}{*}{7} & \multirow{2}{*}{(`Putting [sth] on a surface', `Bottle')} & \multirow{2}{*}{77 / 68.1\%} & (`Putting [sth] upright on the table', `Bottle') (seen) & 33 / 42.9\% & w/ `Bottle' (42.1\%) & w/ `Turning [sth] upside down' (5.3\%) \\
&&& (`Hitting [sth] with [sth]', `Bottle') (seen) & 15 / 19.5\% & w/ `Bottle' (14.8\%) & w/ `Turning [sth] upside down' (5.3\%) \\
\midrule
\multirow{2}{*}{8} & \multirow{2}{*}{(`Showing that [sth] is empty', `Bottle')} & \multirow{2}{*}{70 / 85.4\%} & (`Turning [sth] upside down', `Bottle') (seen) & 13 / 18.6\% & w/ `Bottle' (21.7\%) & w/ `Turning [sth] upside down' (5.3\%) \\
&&& (`Pretending to turn [sth] upside down', `Bottle') (seen) & 8 / 11.4\% & w/ `Bottle' (30.6\%) & w/ `Turning [sth] upside down' (5.3\%) \\
\midrule
\multirow{2}{*}{9} & \multirow{2}{*}{(`Throwing [sth]', `Book')} & \multirow{2}{*}{69 / 84.2\%} & (`[sth] falling like a rock', `Book') (seen) & 16 / 23.2\% & w/ `Bottle' (13.0\%) & w/ `Pushing [sth] from left to right' (4.9\%) \\
&&& (`Throwing [sth]', `Hat') (seen) & 9 / 13.0\% & w/ `Tissue' (7.7\%) & w/ `Throwing [sth]' (35.0\%) \\
\midrule
\multirow{2}{*}{10} & \multirow{2}{*}{(`Closing [sth]', `Drawer')} & \multirow{2}{*}{60 / 69.0\%} & (`Opening [sth]', `Drawer') (seen) & 31 / 51.7\% & w/ `Door' (18.9\%) & w/ `Opening [sth]' (36.8\%) \\
&&& (`Stuffing [sth] into [sth]', `Drawer') (seen) & 6 / 10.0\% & w/ `Box' (19.2\%) & w/ `Opening [sth]' (36.8\%) \\
\bottomrule
\end{tabular}
}
\label{tab:failure_mode_sth_com}
\end{table*}

\begin{table*}[t]
\centering
\caption{\textbf{Top-10 failure cases of the existing ZS-CAR model on unseen compositions of the EK100-com test set.} We employ C2C~\cite{li2024c2c} trained on EK100-com as baseline for analysis. We report the results with unbiased open-world inference scores. 
`Prediction Rate' denotes the proportion of mispredictions classified as the specific composition. 
`Verb/Object-side' indicates the most frequently co-occurring counterpart for each mispredicted component in the training set, along with its co-occurrence ratio.
The top-10 cases listed below account for $\sim$ 39\% of the total mispredictions on unseen compositions.
}

\resizebox{\linewidth}{!}{
\begin{tabular}{l l l l l l l}
\toprule
\multirow{2}{*}{Rank} & \multirow{2}{*}{Ground Truth (Verb, Object)} & Error & \multirow{2}{*}{Top-2 Misprediction (seen/unseen)} & Prediction & \multicolumn{2}{l}{Most Frequent Co-occurring Component} \\ 
\cline{6-7}
&& Count / Rate (\%) && Count / Rate (\%) & Verb-side & Object-side \\
\toprule
\multirow{2}{*}{1} & \multirow{2}{*}{(`put', `knife')} & \multirow{2}{*}{340 / 80.8\%} & (`take', `knife') (seen) & 81 / 23.8\% & w/ `spoon' (6.8\%) & w/ `take' (56.2\%) \\
&&& (`put', `fork') (seen) & 41 / 12.1\% & w/ `spoon' (8.6\%) & w/ `take' (40.3\%) \\
\midrule
\multirow{2}{*}{2} & \multirow{2}{*}{(`put', `lid')} & \multirow{2}{*}{303 / 76.7\%} & (`take', `lid') (seen) & 44 / 14.5\% & w/ `spoon' (6.8\%) & w/ `take' (45.4\%) \\
&&& (`remove', `lid') (seen) & 22 / 7.3\% & w/ `lid' (16.1\%) & w/ `take' (45.4\%) \\
\midrule
\multirow{2}{*}{3} & \multirow{2}{*}{(`take', `plate')} & \multirow{2}{*}{259 / 46.7\%} & (`put', `plate') (seen) & 70 / 27.0\% & w/ `spoon' (8.6\%) & w/ `put' (52.1\%) \\
&&& (`take', `bowl') (seen) & 32 / 12.4\% & w/ `spoon' (6.8\%) & w/ `take' (32.6\%) \\
\midrule
\multirow{2}{*}{4} & \multirow{2}{*}{(`wash', `sink')} & \multirow{2}{*}{143 / 100\%} & (`wash', `top') (seen) & 68 / 47.6\% & w/ `hand' (12.7\%) & w/ `wash' (89.2\%) \\
&&& (`wash', `cloth') (seen) & 14 / 9.8\% & w/ `hand' (12.7\%) & w/ `take' (35.2\%) \\
\midrule
\multirow{2}{*}{5} & \multirow{2}{*}{(`put', `spatula')} & \multirow{2}{*}{131 / 69.0\%} & (`take', `spatula') (seen) & 42 / 32.1\% & w/ `spoon' (6.8\%) & w/ `take' (58.2\%) \\
&&& (`put', `spoon') (seen) & 18 / 13.7\% & w/ `spoon' (6.8\%) & w/ `take' (36.9\%) \\
\midrule
\multirow{2}{*}{6} & \multirow{2}{*}{(`put', `cup')} & \multirow{2}{*}{115 / 56.1\%} & (`take', `cup') (seen) & 25 / 21.7\% & w/ `spoon' (6.8\%) & w/ `take' (48.8\%) \\
&&& (`put', `glass') (seen) & 11 / 9.6\% & w/ `spoon' (8.6\%) & w/ `take' (36.4\%) \\
\midrule
\multirow{2}{*}{7} & \multirow{2}{*}{(`shake', `hand')} & \multirow{2}{*}{94 / 97.9\%} & (`turn-off', `tap') (seen) & 38 / 40.4\% & w/ `tap' (93.6\%) & w/ `turn-on' (49.5\%) \\
&&& (`wash', `hand') (seen) & 18 / 19.1\% & w/ `hand' (12.7\%) & w/ `wash' (59.8\%) \\
\midrule
\multirow{2}{*}{8} & \multirow{2}{*}{(`put', `board or chopping')} & \multirow{2}{*}{89 / 56.0\%} & (`take', `board or chopping') (seen) & 11 / 12.4\% & w/ `spoon' (6.8\%) & w/ `take' (45.1\%) \\
&&& (`put', `knife') (unseen) & 9 / 10.1\% & w/ `spoon' (8.6\%) & w/ `take' (56.2\%) \\
\midrule
\multirow{2}{*}{9} & \multirow{2}{*}{(`wash', `hob')} & \multirow{2}{*}{83 / 70.9\%} & (`wash', `top') (seen) & 38 / 45.8\% & w/ `hand' (12.7\%) & w/ `wash' (89.2\%) \\
&&& (`wash', `oven') (seen) & 7 / 10.3\% & w/ `hand' (12.7\%) & w/ `open' (35.9\%) \\
\midrule
\multirow{2}{*}{10} & \multirow{2}{*}{(`take', `salt')} & \multirow{2}{*}{74 / 70.5\%} & (`put', `salt') (seen) & 9 / 12.2\% & w/ `spoon' (8.6\%) & w/ `put' (38.0\%) \\
&&& (`pour', `salt') (seen) & 7 / 9.5\% & w/ `water' (25.2\%) & w/ `put' (37.9\%) \\

\bottomrule
\end{tabular}
}
\label{tab:failure_mode_epic_com}
\end{table*}

\section{Additional evidence of object-driven shortcuts}
\label{sec:additional_evidence}
In this section, we provide further evidence demonstrating that existing ZS-CAR learning approaches are susceptible to object-driven shortcut learning.

\topic{Failure case analysis of the existing ZS-CAR model.}
In \tabref{failure_mode_sth_com} and \tabref{failure_mode_epic_com}, we investigate top-10 failure cases of the model trained with the state-of-the-art method~\cite{li2024c2c}. 
The top-10 failure cases shown in the \tabref{failure_mode_sth_com} constitute 9.5\% of the misclassifications on unseen compositions for Sth-com~\cite{li2024c2c}, compared to 39\% for EK100-com as shown in \tabref{failure_mode_epic_com}.
For the analysis of Sth-com~\cite{li2024c2c}, we exclude ambiguous cases arising from annotation issues to obtain meaningful results. 
A prime example is the top-1 failure case, (`Covering [something] with [something]', `Paper'), where 66\% of the errors are misclassified as `Pencil'. 
Upon visual inspection, we confirm that nearly all these videos actually depict the action of covering a pencil with paper.
In \tabref{failure_mode_sth_com} and \tabref{failure_mode_epic_com}, we observe a common tendency across both datasets: the model frequently misclassifies verbs as their temporally reversed or opposing counterparts. 
Specifically, we demonstrate that these mispredictions stem from the model's reliance on co-occurrence statistics.
For instance, the model misclassifies unseen `(Closing, Drawer)' samples as `(Opening, Drawer)', which accounts for 36\% of the errors.
Notably, `Opening' is the verb most frequently paired with `Drawer' in the training set. 
These findings suggest that the model has failed to learn robust verb representations that generalize to unseen compositions, largely due to object-driven shortcut learning.

\begin{figure}[t]
\centering
\includegraphics[width=.5\linewidth]{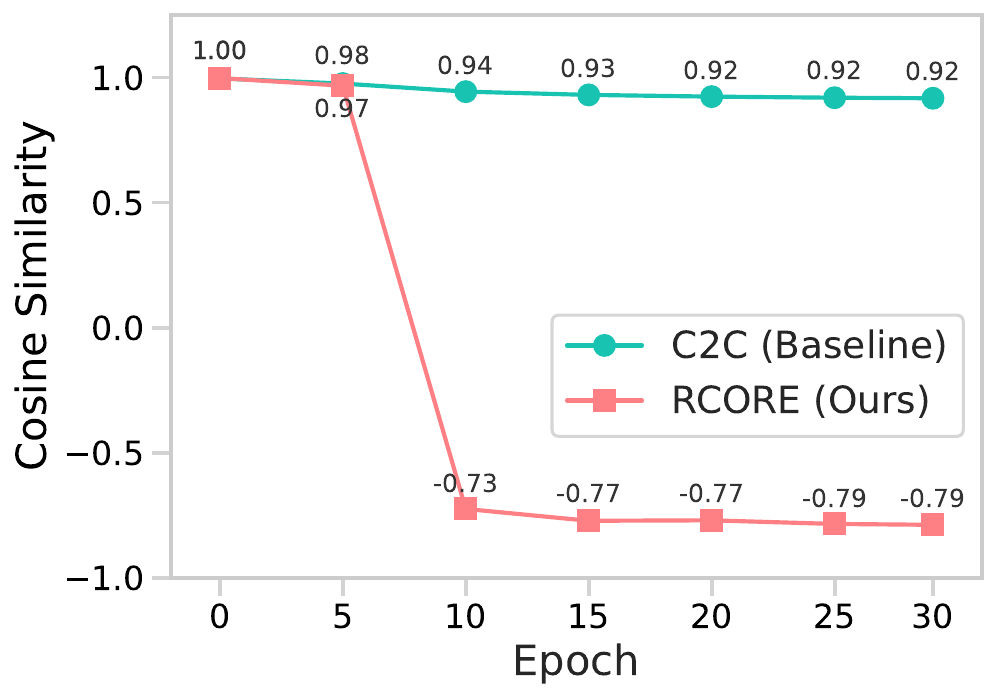}
\vspace{-1em}
\figcapmargin
\figcaption{Similarity analysis between original and reversed verb features}{
    We plot the cosine similarity between the original and reversed verb features for the baseline (C2C~\cite{li2024c2c}) and \ours{} with the CLIP~\cite{radford2021clip} backbone on the Sth-com~\cite{li2024c2c} dataset.
    The baseline maintains a \textit{high similarity} (+0.92) during entire training, revealing limited temporal sensitivity.
    In contrast, the similarity becomes \textit{strongly negative} for \ours{} as training progresses, indicating improved temporal discriminative capability.
}
\label{fig:cosine_sim}
\end{figure}

\topic{Similarity analysis of verb features.}
In \figref{cosine_sim}, we track the cosine similarity between the original verb features $\f^V$ and the reversed verb features $\f^V_{\text{rev}}$ for the baseline~\cite{li2024c2c} over the course of training.
The baseline~\cite{li2024c2c} maintains a high cosine similarity ($+0.92$), revealing its reliance on object-driven shortcuts rather than temporally grounded verb representations.
Additionally, we plot the similarity of \ours{} in \figref{cosine_sim}.
\ours{} drives this similarity to a strongly negative value ($-0.79$), indicating that the model successfully distinguishes opposite temporal semantics, \eg `opening' \vs `closing'.

\section{Additional Results}
\label{sec:additional_results}
\begin{table*}[t]
\centering
\caption{\textbf{Macro Compositional Gap results.} 
    We show Macro Compositional Gap (\mcg{}) results of C2C~\cite{li2024c2c} and \ours{} on the Sth-com~\cite{li2024c2c} and EK100-com datasets. 
    The \textbf{best} numbers are highlighted.
}
\def\arraystretch{1.2}
\resizebox{.5\linewidth}{!}{
\begin{tabular}{ll cc c cc}
\toprule
\multirow{2}{*}{Backbone} & \multirow{2}{*}{Method} & \multicolumn{2}{c}{Sth-com} && \multicolumn{2}{c}{EK100-com} \\ 
\cline{3-4} \cline{6-7} 
&& {Seen} & {Unseen} && {Seen} & {Unseen} \\
\midrule
\multirow{3}{*}{CLIP} 
& AIM~\cite{yang2023aim} & \textbf{+1.18} & $-$0.47 && +0.95 & $-$1.10 \\
& C2C~\cite{li2024c2c} & +0.78 & +0.16 && \textbf{+1.56} & $-$0.38 \\
& \ours{} & +0.53 & \textbf{+0.64} && $-$0.03 & \textbf{+0.11} \\
\midrule
\multirow{3}{*}{\shortstack{InternVideo2}} 
& LoRA~\cite{hu2022lora} & \ 0.00 & $-$0.16 && +0.60 & $-$0.53 \\
& C2C~\cite{li2024c2c} & \textbf{+0.59} & +0.14 && \textbf{+0.83} & $-$0.38 \\
& \ours{} & $-$0.33 & \textbf{+0.76} && +0.02 & \textbf{+0.85} \\
\bottomrule
\end{tabular}
\label{tab:macro_cg}
}
\end{table*}

\subsection{Macro Compositional Gap}
In this section, we define Macro Compositional Gap (\mcg{}).
\mcg{} is the macro version of Compositional Gap (\cg{}),  which averages \cg{} per composition class, ensuring robustness to class imbalance.
Let $\mathbb{Y}_{\text{test}}^{C} \subseteq \mathbb{Y}^V \times \mathbb{Y}^O$ denote the set of \emph{unique} composition labels that appear in the test split.
For each $\y^C=(\y^V,\y^O) \in \mathbb{Y}_{\text{test}}^{C}$, let $\Acc^C_{\y^C}$ be the top-1 composition accuracy restricted to samples of $\y^C$, and let $\Acc^V_{\y^C}$ and $\Acc^O_{\y^C}$ be the corresponding verb/object accuracies on the same samples of $\y^C$, computed from the joint top-1 composition predictions.
Then we define
\begin{equation}
\Delta_{\text{CG}}^{\text{macro}} = \frac{1}{|\mathbb{Y}^C_{\text{test}}|} \sum_{\y^C \in \mathbb{Y}^C_{\text{test}}} \left[ \text{Acc}^C_{\y^C} - \left( \text{Acc}^V_{\y^C} \times \text{Acc}^O_{\y^C} \right) \right].
\label{eq:mcg}
\end{equation}
In \tabref{macro_cg}, we report \mcg{} performances of baselines and \ours{} on the Sth-com~\cite{li2024c2c} and EK100-com datasets.
\ours{} consistently achieves the highest \mcg{} on unseen compositions across different backbones and datasets, demonstrating its robust compositional reasoning capabilities.

\subsection{Shortcut diagnosis on another ZS-CAR method}
\begin{figure}[t]
\centering  
\includegraphics[width=.5\linewidth]{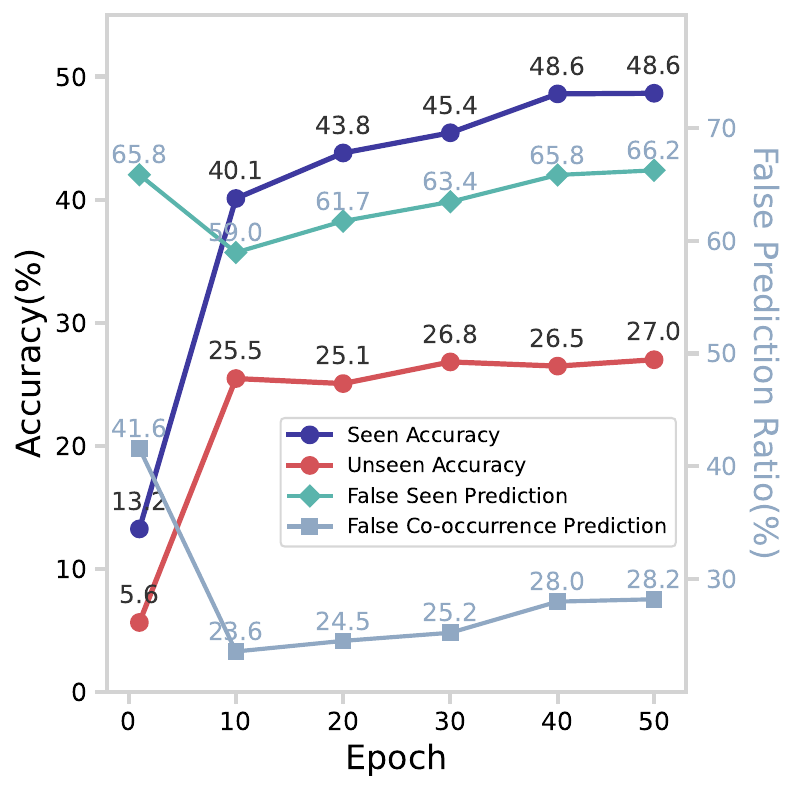}
\\
\figcapmargin
\figcaption{Learning curve of Jung et al. on Sth-com}{
We report the learning curve of Jung et al.~\cite{jung2025crr} along with two training-bias metrics, FSP and FCP.
Even with its enhanced disentangling training strategy, the seen-unseen accuracy gap grows together with FSP/FCP, indicating strong overfitting to seen compositions.
}
\label{fig:crr_learning_curve}
\end{figure}

We apply the FSP/FCP diagnosis to another recent ZS-CAR method, Jung et al.~\cite{jung2025crr}, to test whether shortcut reliance is specific to C2C~\cite{li2024c2c} or generalizes across ZS-CAR methods.
As shown in \figref{crr_learning_curve}, even with an enhanced disentangling training strategy, Jung et al.~\cite{jung2025crr} with the CLIP backbone~\cite{radford2021clip} also suffers from severe overfitting to co-occurrence statistics: the seen--unseen accuracy gap grows together with FSP/FCP, exhibiting the same trend observed for C2C~\cite{li2024c2c}.
This indicates that shortcut reliance is not specific to a single baseline but is a common failure mode across ZS-CAR methods.

\subsection{Additional ablation results}
In \tabref{supple_additional_ablation}, we provide the comprehensive ablation study results to examine each design choice of \ours{}, using the Sth-com~\cite{li2024c2c} dataset. 

\topic{Effects of \torc{} with different temporal aggregation methods.}
In \tabref{supple_additional_ablation} (a), we demonstrate that the weak temporal order modeling caused by object-driven shortcut learning cannot be resolved merely by employing more sophisticated temporal aggregation methods. 
We compare the baseline's temporal average pooling with attention pooling (Attn. pool). In attention pooling, a linear layer computes frame-wise attention weights from the input verb feature sequences, aggregating them via a weighted sum to obtain the final verb feature for classification. 
While attention pooling yields a slight improvement in verb accuracy, it fails to reduce the compositional gap ($\Delta_{CG}$) on unseen compositions. 
In contrast, our \torc{} achieves significant improvements in harmonic mean (H.M.) with both aggregation methods. 
Notably, the combination of temporal average pooling and \torc{} enables the model to overcome the negative compositional gap.
\begin{table*}[t]
\centering

\caption{\textbf{Additional ablation study of \ours{} on Sth-com.} 
    We provide the additional ablation study results on Sth-com~\cite{li2024c2c}.
    In every experiment, we use C2C~\cite{li2024c2c} with CLIP-B/16~\cite{radford2021clip} as our baseline. 
    We report performance on both seen and unseen compositions, along with their harmonic mean (H.M.).
    The \textbf{best} numbers are highlighted.
}

\mpage{0.55}{\scriptsize(a) Effects of \torc{} with different\\temporal aggregation.}
\mpage{0.42}{\scriptsize(b) Effects of input samples in \cpr{}.}
\\
\mpage{0.55}{
    \resizebox{\linewidth}{!}{
        \begin{tabular}{l c cc c cc c ccc}
        \toprule
        \multirow{2}{*}{Method} &
        \multirow{2}{*}{\torc} & \multicolumn{2}{c}{Verb} && \multicolumn{2}{c}{Object} && \multicolumn{3}{c}{Composition} \\ 
        \cline{3-4} \cline{6-7} \cline{9-11} 
        && V@S & V@U && O@S & O@U && Seen ($\Dcg$) & Unseen ($\Dcg$) & H.M. \\
        \midrule
        Average & $\times$ & 63.60 & 54.36 && 67.72 & 56.10 && 46.31 (+3.24) & 30.08 (-0.42) & 36.47 \\
        Attn. pool & $\times$ & 64.28 & 54.76 && 67.50 & 55.73 && 47.11 (+3.72) & 30.09 (-0.43) & 36.72 \\
        Average & \checkmark & 65.65 & 56.80 && 68.59 & 55.19 && 48.65 (+3.62) & 31.35 (-0.00) & \textbf{38.13} \\
        Attn. pool & \checkmark & 65.51 & 56.50 && 67.80 & 55.00 && 48.07 (+3.65) & 30.78 (-0.30) & 37.53 \\
        \bottomrule
        \end{tabular}
    }
}
\hfill
\mpage{0.42}{
\resizebox{\linewidth}{!}{
\begin{tabular}{l cc cc ccc}
\toprule
\multirow{2}{*}{Inputs} & \multicolumn{2}{c}{Verb} & \multicolumn{2}{c}{Object} & \multicolumn{3}{c}{Composition} \\
\cline{2-3} \cline{4-5} \cline{6-8}
& V@S & V@U & O@S & O@U & Seen & Unseen & H.M. \\
\midrule
Original & 62.54 & 52.97 & 56.90 & 57.05 & 37.43 & 31.41 & 34.16 \\
Mixup~\cite{zhang2018mixup} & 62.37 & 52.56 & 63.50 & 59.20 & 41.62 & 31.45 & 35.83 \\
\cpr{} w/o F.E.~\cite{ding2022fame} & 63.23 & 54.29 & 65.15 & 56.49 & 43.57 & 30.86 & 36.13 \\
\cpr{} w/ F.E.~\cite{ding2022fame} & 64.17 & 54.55 & 63.60 & 58.70 & 42.74 & 33.05 & \textbf{37.28} \\
\bottomrule
\end{tabular}
\label{tab:ablation_input_cpr}
}
}
\\
\vspace{.2em}
\mpage{0.42}{
{\scriptsize(c) Effects of hyperparameters of \cpr{}.}
\centering
\resizebox{.9\linewidth}{!}{
\begin{tabular}{l cc c cc c ccc}
\toprule

\multirow{2}{*}{$p_\text{CPR}$} 
& \multicolumn{2}{c}{Verb} && \multicolumn{2}{c}{Object} && \multicolumn{3}{c}{Composition} \\ 
\cline{2-3} \cline{5-6} \cline{8-10} 
& V@S & V@U && O@S & O@U && Seen & Unseen & H.M. \\
\midrule
0.8 & 64.66 & 52.89 && 65.06 & 58.77 && 45.20 & 31.14 & 36.88 \\
0.9 & 64.70 & 55.50 && 64.73 & 57.69 && 44.75 & 32.60 & \textbf{37.72} \\
0.95 & 63.95 & 54.05 && 63.97 & 59.52 && 43.09 & 33.22 & 37.51 \\
1.0 & 64.95 & 55.50 && 61.49 & 57.20 && 41.82 & 32.84 & 36.79 \\
\bottomrule
\end{tabular}
}
}
\mpage{0.55}{
{\scriptsize(d) Effects of hyperparameters of \torc{}.}
\resizebox{\linewidth}{!}{
\begin{tabular}{l cc cc c cc c ccc}
\toprule

\multirow{2}{*}{$\gamma$} & scaling & scaling & \multicolumn{2}{c}{Verb} && \multicolumn{2}{c}{Object} && \multicolumn{3}{c}{Composition} \\ 
\cline{4-5} \cline{7-8} \cline{10-12} 
& start epoch & end epoch & V@S & V@U && O@S & O@U && Seen & Unseen & H.M. \\
\midrule
0.5 & 1 & 1 & 65.59 & 55.63 && 68.49 & 54.80 && 48.33 & 30.19 & 37.17 \\
1.0 & 1 & 1 & 64.88 & 55.81 && 67.21 & 55.12 && 47.06 & 30.69 & 37.15 \\
1.0 & 5 & 5 & 64.93 & 56.03 && 67.67 & 55.81 && 47.20 & 31.38 & 37.70 \\
1.0 & 5 & 10 & 65.65 & 56.80 && 68.59 & 55.19 && 48.65 & 31.35 & \textbf{38.13} \\
\bottomrule
\end{tabular}
}
\\
\vspace{.5em}
}
\mpage{0.6}{
{\scriptsize(e) Effects of hyperparameters of $L_\text{CPR}$.}
\resizebox{\linewidth}{!}{
\begin{tabular}{cc cc cc c cc c ccc}
\toprule

\multirow{2}{*}{$m$} & \multirow{2}{*}{$\delta$} & scaling & scaling & \multicolumn{2}{c}{Verb} && \multicolumn{2}{c}{Object} && \multicolumn{3}{c}{Composition} \\ 
\cline{5-6} \cline{8-9} \cline{11-13} 
&& start epoch & end epoch & V@S & V@U && O@S & O@U && Seen & Unseen & H.M. \\
\midrule
0.5 & 0.1 & 15 & 15 & 67.06 & 58.13 && 63.67 & 58.66 && 45.00 & 35.16 & \textbf{39.48} \\
0.5 & 0.1 & 15 & 20 & 66.16 & 58.78 && 63.75 & 57.96 && 44.65 & 34.89 & 39.17 \\
0.5 & 0.5 & 15 & 15 & 66.16 & 57.03 && 62.75 & 57.56 && 43.55 & 33.77 & 38.04 \\
0.5 & 0.5 & 15 & 20 & 67.04 & 56.77 && 61.74 & 57.56 && 43.37 & 34.30 & 38.31 \\
1.0 & 0.1 & 15 & 15 & 66.31 & 58.56 && 63.77 & 57.99 && 44.47 & 34.64 & 38.94 \\
1.0 & 0.1 & 15 & 20 & 66.88 & 58.24 && 63.88 & 57.01 && 44.90 & 34.21 & 38.83 \\
\bottomrule
\end{tabular}
}
}

\label{tab:supple_additional_ablation}
\end{table*}

\topic{Effects of input samples in \cpr{}.}
In \tabref{supple_additional_ablation} (b), we validate the effects of input sample types in \cpr{}.
We argue that conventional image-based augmentations are insufficient for representing newly created unseen composition labels. 
For example, Mixup~\cite{zhang2018mixup} blends verbs and objects concurrently, failing to preserve the temporal difference information that is essential for verb recognition.
As a result, these methods yield a lower verb@unseen-comp than the baseline (52.6\% vs. 53.0\%).
In contrast, our \cpr{} adopts a static frame mixing strategy, aiming to generate novel compositions by mixing only object information while preserving the verbs, shows modest performance gains over the baseline for both verb  and object accuracy on unseen compositions, resulting in the best H.M. 
Furthermore, we observe that mixing a static frame into the estimated foreground region shows additional  performance boost (+1.2 points in the H.M.) over applying it to the entire background, as it creates more realistic samples.
We denote the latter as `\cpr{} w/o F.E.', indicating `without foreground estimation'.

\begin{table*}[t]
\centering

\caption{\textbf{Ablation study on EK100-com.} 
    We provide the results of the ablation study on EK100-com presented in the main paper.
    In every experiment, we use C2C~\cite{li2024c2c} with CLIP-B/16~\cite{radford2021clip} as our baseline. 
    We report performance on both seen and unseen compositions, along with their harmonic mean (H.M.).
}
\mpage{0.48}{\scriptsize
(a) Effects of \cpr{} and TORC.
\resizebox{\linewidth}{!}{
\begin{tabular}{cc cc c cc c ccc}
\toprule

\multirow{2}{*}{\cpr{}} & \multirow{2}{*}{\torc{}} 
& \multicolumn{2}{c}{Verb} && \multicolumn{2}{c}{Object} && \multicolumn{3}{c}{Composition)} \\ 
\cline{3-4} \cline{6-7} \cline{9-11} 
&& V@S & V@U && O@S & O@U && Seen & Unseen & H.M. \\
\midrule
&& 66.19 & 49.71 && 56.61 & 47.48 && 42.72 & 22.38 & 29.38 \\
\checkmark & & 66.06 & 51.28 && 54.23 & 50.56 && 38.93 & 27.10 & 31.96 \\
& \checkmark & 65.90 & 52.31 && 57.28 & 46.74 && 43.48 & 23.67 & 30.65 \\
\checkmark & \checkmark & 67.52 & 54.08 && 54.00 & 49.04 && 40.13 & 27.40 & \textbf{32.56} \\
\bottomrule
\end{tabular}
\label{tab:ek100_ablation_component}
}
}
\mpage{0.48}{\scriptsize
(b) Effects of each \torc{} loss term.
\resizebox{\linewidth}{!}{
\begin{tabular}{cc cc c cc c ccc}
\toprule

\multirow{2}{*}{$L_{\cos}$} & \multirow{2}{*}{$L_\text{ent}$} 
& \multicolumn{2}{c}{Verb} && \multicolumn{2}{c}{Object} && \multicolumn{3}{c}{Composition} \\ 
\cline{3-4} \cline{6-7} \cline{9-11} 
&& V@S & V@U && O@S & O@U && Seen & Unseen & H.M. \\
\midrule
&& 66.19 & 49.71 && 56.61 & 47.48 && 42.72 & 22.38 & 29.38 \\
\checkmark & & 65.43 & 52.11 && 56.54 & 47.44 && 42.67 & 24.24 & \textbf{30.92} \\
& \checkmark & 66.00 & 51.06 && 56.64 & 47.37 && 42.82 & 22.53 & 29.53 \\
\checkmark & \checkmark & 65.90 & 52.31 && 57.28 & 46.74 && 43.48 & 23.67 & 30.65 \\
\bottomrule
\end{tabular}
}
}
\mpage{0.48}{
\scriptsize (c) Effects of $L_{\text{CPR}}$ in \cpr{}.
\resizebox{\linewidth}{!}{
\begin{tabular}{c cc c cc c ccc}
\toprule

\multirow{2}{*}{$L_\text{CPR}$}
& \multicolumn{2}{c}{Verb} && \multicolumn{2}{c}{Object} && \multicolumn{3}{c}{Composition} \\ 
\cline{2-3} \cline{5-6} \cline{8-10} 
& V@S & V@U && O@S & O@U && Seen & Unseen & H.M. \\
\midrule
& 66.35 & 50.79 && 54.72 & 50.16 && 39.25 & 26.60 & 31.71 \\
\checkmark & 66.06 & 51.28 && 54.23 & 50.56 && 38.93 & 27.10 & \textbf{31.96} \\
\bottomrule
\end{tabular}
}
}
\label{tab:ek100_ablation_torc}
\end{table*}

\topic{Effects of hyperparameters of \ours{}.}
In \tabref{supple_additional_ablation} (c), we show the effect of $p_\text{CPR}$, the per-batch probability for CPR application, revealing a trade-off between seen and unseen composition accuracies. 
Decreasing $p_\text{CPR}$ slightly from 1.0 (\eg 0.95 or 0.9) yields optimal unseen accuracy improvements with minimal drops in seen performance, whereas lowering it further (\eg $p_\text{CPR} = 0.8$) degrades the gains on unseen pairs. 
Consequently, we tune $p_\text{CPR}$ for each dataset and backbone based on the best validation H.M. of seen and unseen accuracies, as shown in \tabref{hyperparam}.
In \tabref{supple_additional_ablation} (d), we observe that introducing the explicit temporal order modeling task via \torc{} starting from 5 epoch, after initial training, is effective.
In particular, the strategy of linearly increasing the loss weight from epoch 5 to 10 is the most effective. %
Finally, in \tabref{supple_additional_ablation} (e), we analyze the impact of the loss weights of $L_\text{CPR}$ in \cpr{}. Increasing the margin $m$ or loss weight $\delta$ degrades performance, suggesting that excessive training penalties are detrimental. 
We find $m=0.5$ and $\delta=0.1$ to be optimal. Contrary to the results in \tabref{supple_additional_ablation} (d), maintaining a constant low loss weight ($\delta=0.1$) for $L_\text{CPR}$ yields better performance than applying progressive weight scaling.

\subsection{Ablation studies on EK100-com}
In \tabref{ek100_ablation_component}, we provide the ablation study results conducted on the EK100-com dataset, complementing the results discussed with the Sth-com~\cite{li2024c2c} dataset in the main paper. 
In \tabref{ek100_ablation_component} (a), we demonstrate that while \cpr{} and \torc{} are individually effective, integrating both components yields the best performance on EK100-com, outperforming the baseline in H.M. by 3.2 points.
In \tabref{ek100_ablation_component} (b), using only $L_\text{cos}$ in \torc{} yields the highest H.M. on EK100-com.
In \tabref{ek100_ablation_component} (c), surpressing co-occurrence priors with $L_\text{CPR}$ improves H.M. by 0.3 points, consistent with the results on Sth-com~\cite{li2024c2c}.

\subsection{Results with validation-set-tuned bias calibration}
\begin{table*}[t]
\centering
\caption{\textbf{Sth-com results with bias calibration.} 
    We show the top-1 verb, object, and composition classification accuracies (\%) on the Sth-com~\cite{li2024c2c} test set. 
    We tune the bias term for unseen composition predictions on the validation set and apply the selected bias (`Best Bias') to the test set. 
    We report performance on both seen and unseen compositions, along with their harmonic mean (H.M.).
    The \textbf{best} numbers are highlighted.
}
\def\arraystretch{1.2}
\resizebox{\linewidth}{!}{
\begin{tabular}{ll cc ccc c ccc c ccc}
\toprule
\multirow{3}{*}{Backbone} & \multirow{3}{*}{Method} & \cellcolor{gray!15}
&& \multicolumn{3}{c}{Verb} && \multicolumn{3}{c}{Object} && \multicolumn{3}{c}{Composition} \\ 
\cline{5-7} \cline{9-11} \cline{13-15} 
&& \cellcolor{gray!15} && \multirow{2}{*}{\shortstack{\rule{0pt}{2.8ex}@Seen\\Comp}} & \multirow{2}{*}{\shortstack{\rule{0pt}{2.8ex}@Unseen\\Comp}} & \multirow{2}{*}{H.M.} && \multirow{2}{*}{\shortstack{\rule{0pt}{2.8ex}@Seen\\Comp}} & \multirow{2}{*}{\shortstack{\rule{0pt}{2.8ex}@Unseen\\Comp}} & \multirow{2}{*}{H.M.} && \multirow{2}{*}{Seen (\cg{})} & \multirow{2}{*}{Unseen (\cg{})} & \multirow{2}{*}{H.M.} \\
&& \cellcolor{gray!15}\multirow{-3}{*}{\shortstack{Best\\Bias}}
\\
\midrule
\multirow{3}{*}{CLIP} & AIM~\cite{yang2023aim} & \cellcolor{gray!15}+0.01 && 50.32 & 43.19 & 46.48 && 64.02 & 54.78 & 59.04 && 33.40 (+1.19) & 24.60 (+0.94) & 28.33 \\
& C2C~\cite{li2024c2c} & \cellcolor{gray!15}+0.07 && 60.97 & 56.62 & 58.71 && 65.05 & 56.98 & 60.75 && 42.06 (+2.40) & 33.98 (+1.72) & 37.59 \\
& \ours{} & \cellcolor{gray!15}\ \ 0.00 && 65.73 & 59.00 & \textbf{62.18} && 64.79 & 56.34 & 60.27 && \textbf{44.99 (+2.40)} & \textbf{33.90 (+0.66)} & \textbf{38.67} \\
\midrule
\multirow{3}{*}{\shortstack{InternVideo2}} & LoRA~\cite{hu2022lora} & \cellcolor{gray!15}+0.02 && 42.02 & 40.74 & 41.37 && 64.04 & 59.75 & 61.82 && 26.00 ($-$0.91) & 25.09 (+0.75) & 25.54 \\
& C2C~\cite{li2024c2c} & \cellcolor{gray!15}+0.03 && 68.54 & 65.27 & 66.87 && 68.58 & 64.58 & \textbf{66.52} && 48.42 (+1.41) & 42.66 (+0.51) & 45.36 \\
& \ours{} & \cellcolor{gray!15}\ \ 0.00 && 71.65 & 66.65 & \textbf{69.06} && 67.96 & 64.56 & 66.22 && \textbf{50.20 (+1.51)} & \textbf{43.98 (+0.95)} & \textbf{46.88} \\
\bottomrule
\end{tabular}
}
\label{tab:biased_sth_results}
\end{table*}

\vspace{-.5em}

In this section, we provide the open-world biased results with validation-set-tuned calibration.
\textit{Bias calibration} is a standard technique in compositional zero-shot learning~\cite{purushwalkam2019taskdriven,mancini2021compcos,kim2023hierarchical,nayak2023csp, wu2025cond, li2024context} and ZS-CAR~\cite{li2024c2c,jiang2025dhd,jung2025crr,ye2025logic}. 
It compensates for the inherent issue where logits for unseen compositions---not optimized during training---are naturally lower than those for seen compositions. 
Specifically, it adds a scalar bias term exclusively to the unseen composition logits. 
Previous ZS-CAR works~\cite{li2024c2c,jiang2025dhd,jung2025crr,ye2025logic} often tuned this bias using test-set ground truth under a closed-world setting to report the `best' seen and unseen accuracies. 
We argue that this approach is highly unrealistic.

Instead, we adopt an open-world \textit{validation-set-tuned} bias calibration. 
We tune the bias on the validation set using open-world logits and then apply it to the test set. 
This approach remains applicable to real-world scenarios while effectively improving unseen composition performance as a post-processing step.

In \tabref{biased_sth_results}, we present the calibrated results of the baselines and \ours{} on the Sth-com~\cite{li2024c2c} dataset. 
Notably, the calibrated performance of \ours{} is \textit{identical} to its original unbiased performance. 
This indicates that our training label space expansion strategy via our proposed \cpr{} is effective.
Unlike baseline methods that overfit solely to seen compositions, it successfully provides meaningful supervision for unseen compositions during training.
Furthermore, the \textit{unbiased} performance of \ours{} consistently outperforms the best calibrated performance of all baselines. 
This demonstrates that \ours{} is highly suitable for real-world applications, delivering strong ZS-CAR performance without the need for any post-processing.

\subsection{Results with large video-pretrained VLM backbone}
\label{sec:iv2_1b_results}
\begin{table*}[t]
\centering
\caption{\textbf{Sth-com results with InternVideo2-1B.} 
    We show the top-1 verb, object, and composition classification accuracies (\%) on the Sth-com~\cite{li2024c2c} and EK100-com datasets with the InternVideo2-1B~\cite{wang2024iv2} backbone. 
    The \textbf{best} numbers are highlighted.
}
\vspace{-1em}
{\scriptsize Sth-com~\cite{li2024c2c} results}
\\
\vspace{.5em}
\def\arraystretch{1.2}
\resizebox{.85\linewidth}{!}{
\begin{tabular}{l c ccc c ccc c ccc}
\toprule
\multirow{3}{*}{Method}
&& \multicolumn{3}{c}{Verb} && \multicolumn{3}{c}{Object} && \multicolumn{3}{c}{Composition} \\ 
\cline{3-5} \cline{7-9} \cline{11-13} 
&& \multirow{2}{*}{\shortstack{\rule{0pt}{2.8ex}@Seen\\Comp}} & \multirow{2}{*}{\shortstack{\rule{0pt}{2.8ex}@Unseen\\Comp}} & \multirow{2}{*}{H.M.} && \multirow{2}{*}{\shortstack{\rule{0pt}{2.8ex}@Seen\\Comp}} & \multirow{2}{*}{\shortstack{\rule{0pt}{2.8ex}@Unseen\\Comp}} & \multirow{2}{*}{H.M.} && \multirow{2}{*}{Seen (\cg{})} & \multirow{2}{*}{Unseen (\cg{})} & \multirow{2}{*}{H.M.} \\
\\
\midrule
C2C~\cite{li2024c2c} && 67.50 & 58.21 & 62.51 && 70.58 & 61.62 & 65.80 && 50.23 (+2.59) & 34.96 ($-$0.91) & 41.23 \\
\ours{} && 67.49 & 60.47 & 63.79 && 68.83 & 63.63 & 66.13 && 48.27 (+1.82) & 38.35 ($-$0.13) & \textbf{42.74} \\
\bottomrule
\end{tabular}
\label{tab:iv2_1b_sth_results}
}
\vspace{.5em}
\\
{\scriptsize EK100-com results}
\\
\vspace{.5em}
\def\arraystretch{1.2}
\resizebox{.85\linewidth}{!}{
\begin{tabular}{l c ccc c ccc c ccc}
\toprule
\multirow{3}{*}{Method}
&& \multicolumn{3}{c}{Verb} && \multicolumn{3}{c}{Object} && \multicolumn{3}{c}{Composition} \\ 
\cline{3-5} \cline{7-9} \cline{11-13} 
&& \multirow{2}{*}{\shortstack{\rule{0pt}{2.8ex}@Seen\\Comp}} & \multirow{2}{*}{\shortstack{\rule{0pt}{2.8ex}@Unseen\\Comp}} & \multirow{2}{*}{H.M.} && \multirow{2}{*}{\shortstack{\rule{0pt}{2.8ex}@Seen\\Comp}} & \multirow{2}{*}{\shortstack{\rule{0pt}{2.8ex}@Unseen\\Comp}} & \multirow{2}{*}{H.M.} && \multirow{2}{*}{Seen (\cg{})} & \multirow{2}{*}{Unseen (\cg{})} & \multirow{2}{*}{H.M.} \\
\\
\midrule
C2C~\cite{li2024c2c} && 66.95 & 52.22 & 58.67 && 60.57 & 50.81 & 55.26 && 45.91 (+5.36) & 26.66 (+0.13) & 33.73 \\
\ours{} && 66.36 & 52.88 & 58.86 && 56.83 & 52.10 & 54.36 && 41.86 (+4.16) & 28.79 (+1.24) & \textbf{34.12} \\
\bottomrule
\end{tabular}
\label{tab:iv2_1b_results}
}
\end{table*}

\begin{figure*}[t]
\centering
\mpage{0.45}{
\includegraphics[width=\linewidth]{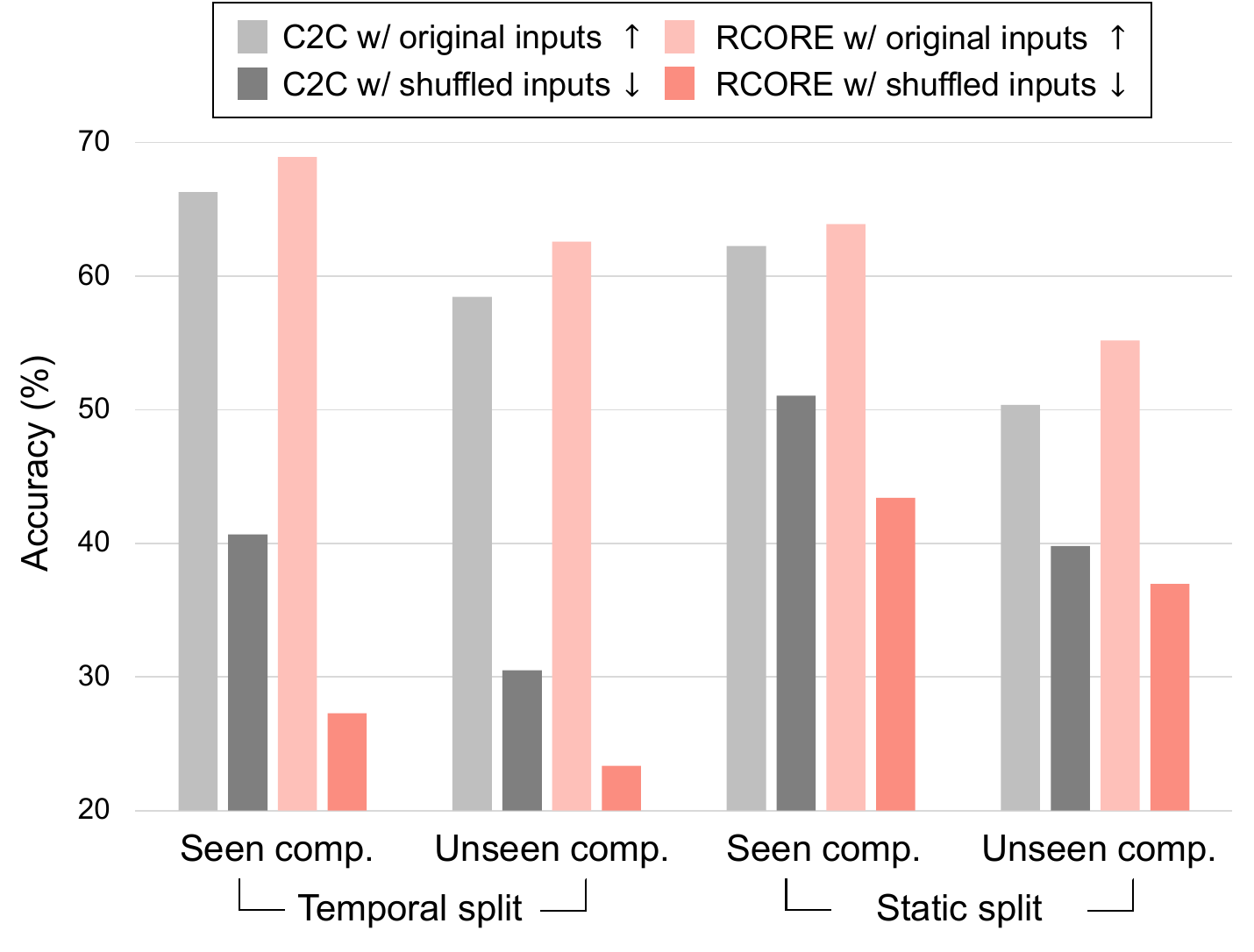}}
\mpage{0.45}{
\includegraphics[width=\linewidth]{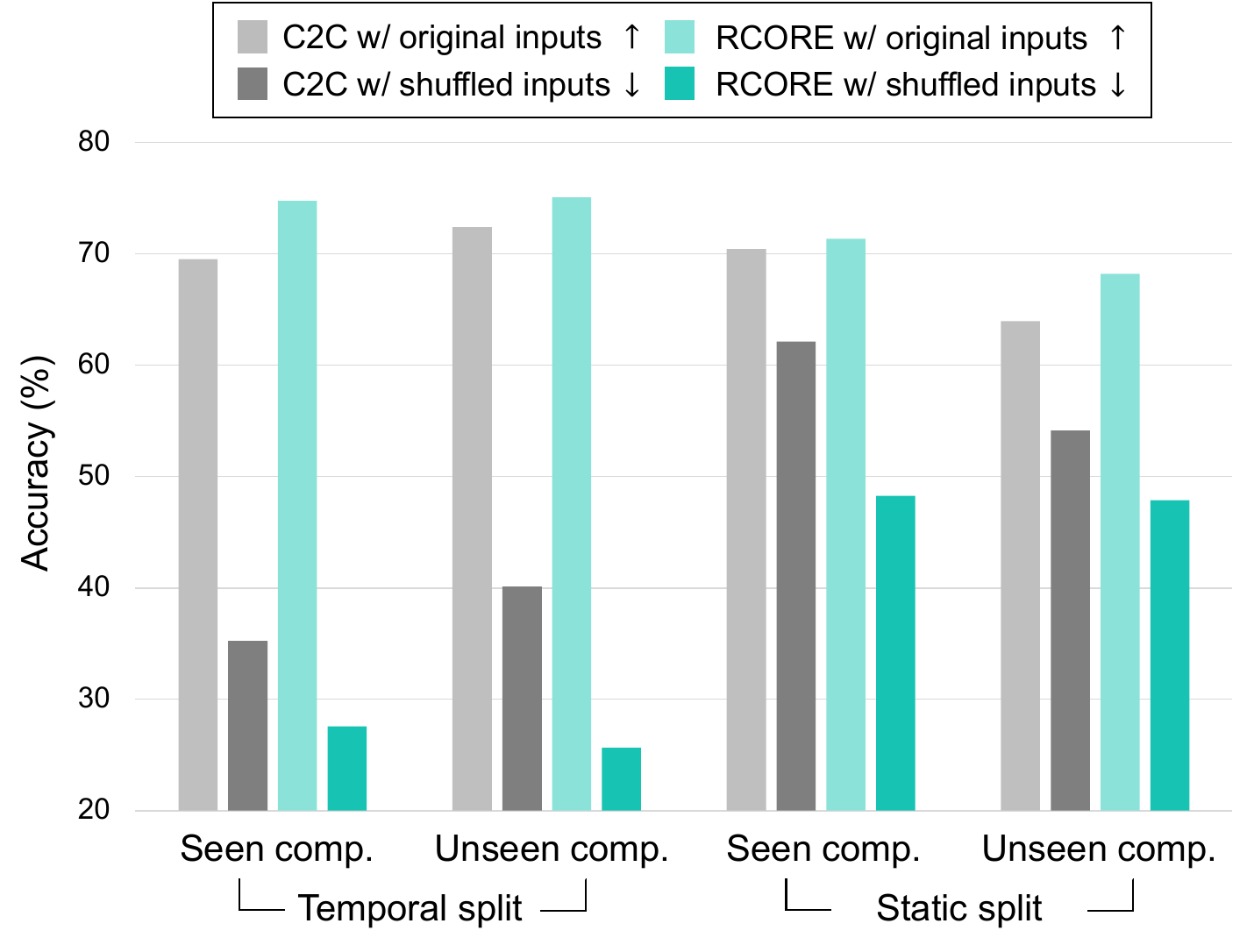}}
\\
\mpage{0.45}{\scriptsize
(a) Our reconstructed Temporal/Static splits}
\mpage{0.45}{
\scriptsize (b) Temporal/Static splits from Sevilla et al.~\cite{laura2021temporal}}
\\
\figcapmargin
\figcaption{Performances on Temporal/Static split of Sth-com}{We evaluate the models on Sth-com~\cite{li2024c2c} using both (a) our reconstructed splits and (b) the splits from Sevilla et al~\cite{laura2021temporal}. 
We utilize both original and temporally shuffled inputs to assess the model’s temporal modeling capability and its reliance on static cues. 
A larger performance gap between original and shuffled inputs indicates that the model predicts verbs more based on temporal dynamics rather than on static cues.}
\label{fig:temporal_static_full}
\end{figure*}

We compare \ours{} and the baseline (C2C~\cite{li2024c2c}) on Sth-com~\cite{li2024c2c} and EK100-com using the InternVideo2-1B~\cite{wang2024iv2} backbone, whose 1B-parameter vision encoder is approximately $11\times$ larger than the 87M-parameter vision encoders used in the main paper.
In \tabref{iv2_1b_sth_results}, \ours{} yields consistent improvements over the baseline on Sth-com~\cite{li2024c2c} and EK100-com, boosting unseen composition accuracy (+3.4p., +2.1p.), unseen \cg{} (+0.8p., +1.1p.), and H.M. (+1.5p., +0.4p.), respectively.
These results demonstrate that its powerful representation learning capability is not constrained by the backbone's parameter size.

\subsection{Temporal/Static split of Sth-com}
We present extended results corresponding to the experiment shown in Figure 5 (c) of the main paper, including all versions of the Temporal/Static split. 
Inspired by prior work~\cite{laura2021temporal, yun2022time}, we rigorously evaluate the temporal modeling capabilities by partitioning the Sth-com~\cite{li2024c2c} dataset into Temporal and Static splits, as shown in \figref{temporal_static_full}.
We evaluate both the baseline (C2C~\cite{li2024c2c}) and \ours{} on the two types of splits, (i) our reconstructed splits and (ii) the splits proposed by Sevilla et al.~\cite{laura2021temporal}.
In all cases, \ours{} exhibits lower performance on shuffled inputs but outperforms the baseline on original inputs. 
This indicates that \ours{} predicts verbs without relying on static cues, leading to robust improvements in verb performance regardless of the split.

In this paragraph, we describe the differences between two Temporal/Static splits.
Sevilla et al.~\cite{laura2021temporal} originally defined these splits based on cognitive science experiments on Something-Something V2~\cite{goyal2017something} verb classes: a Temporal split (18 verbs) where temporal information matters, and a Static split (18 verbs) where such information is redundant. 
However, noting that this Static split includes verbs requiring complex temporal reasoning such as `Folding', we reconfigure the splits based on verb semantics. 
Specifically, we categorized verbs based on this criterion: those requiring temporal information for discrimination are assigned to the Temporal split (62 verbs), while verbs distinguishable by momentary keyframes are assigned to the Static split (80 verbs).
The indices of the verbs in the Temporal split are as follows: 0, 1, 5, 6, 14, 25, 26, 29, 30, 32, 33, 34, 35, 38, 39, 40, 41, 42, 43, 44, 45, 46, 58, 59, 60, 61, 65, 66, 67, 68, 72, 73, 74, 75, 76, 78, 79, 85, 86, 90, 91, 92, 98, 100, 103, 109, 122, 123, 131, 134, 136, 137, 139, 140, 145, 146, 147, 148, 149, 150, 153, 154.
The indices of the verbs in the Static split are as follows: 2, 3, 8, 9, 10, 11, 12, 13, 15, 16, 17, 18, 19, 20, 21, 22, 23, 24, 27, 28, 31, 36, 37, 47, 48, 49, 50, 51, 52, 53, 54, 55, 77, 80, 81, 87, 88, 93, 94, 95, 96, 97, 99, 101, 102, 104, 105, 106, 110, 111, 112, 113, 114, 115, 116, 117, 118, 119, 120, 121, 126, 127, 128, 129, 130, 132, 133, 135, 138, 141, 142, 143, 151, 152, 155, 156, 157, 158, 159, 160.
To balance the sample sizes between the Temporal and Static splits, we excluded a few verbs with the fewest samples from the Temporal split.

\end{document}